\documentclass[twocolumn,letterpaper]{IEEEAerospaceCLS}  

\usepackage[]{graphicx}    
\newcommand{\ignore}[1]{}  

\usepackage{algorithm}
\usepackage[noend]{algpseudocode}
\tabularnewline

\usepackage{amsmath}
\usepackage{amssymb}
\usepackage{booktabs}

\usepackage[caption=false,font=footnotesize]{subfig}
\usepackage{fixltx2e}

\begin{document}
\title{Pose Estimation for Non-Cooperative Spacecraft Rendezvous Using Convolutional Neural Networks}

\author{%
Sumant Sharma\\
Ph.D. Candidate\\ 
Department of Aeronautics \& Astronautics\\
Stanford University\\
496 Lomita Mall, Stanford CA 94305\\
sharmas@stanford.edu
\and 
Connor Beierle\\
Ph.D. Candidate\\ 
Department of Aeronautics \& Astronautics\\
Stanford University\\
496 Lomita Mall, Stanford CA 94305\\
cbeierle@stanford.edu
\and
Simone D'Amico\\
Assistant Professor\\ 
Department of Aeronautics \& Astronautics\\
Stanford University\\
496 Lomita Mall, Stanford CA 94305\\
damicos@stanford.edu
\thanks{\footnotesize 978-1-5386-2014-4/18/$\$31.00$ \copyright2018 IEEE}              
}

\maketitle

\thispagestyle{plain}
\pagestyle{plain}

\begin{abstract}
On-board estimation of the pose of an uncooperative target spacecraft is an essential task for future on-orbit servicing and close-proximity formation flying missions. However, two issues hinder reliable on-board monocular vision based pose estimation: robustness to illumination conditions due to a lack of reliable visual features and scarcity of image datasets required for training and benchmarking. To address these two issues, this work details the design and validation of a monocular vision based pose determination architecture for spaceborne applications. The primary contribution to the state-of-the-art of this work is the introduction of a novel pose determination method based on Convolutional Neural Networks (CNN) to provide an initial guess of the pose in real-time on-board. The method involves discretizing the pose space and training the CNN with images corresponding to the resulting pose labels. 
Since reliable training of the CNN requires massive image datasets and computational resources, the parameters of the CNN must be determined prior to the mission with synthetic imagery. Moreover, reliable training of the CNN requires datasets that appropriately account for noise, color, and illumination characteristics expected in orbit. Therefore, the secondary contribution of this work is the introduction of an image synthesis pipeline, which is tailored to generate high fidelity images of any spacecraft 3D model. In contrast to prior techniques demonstrated for close-range pose determination of spacecraft, the proposed architecture relies on neither hand-engineered image features nor a-priori relative state information. Hence, the proposed technique is scalable to spacecraft of different structural and physical properties as well as robust to the dynamic illumination conditions of space. Through metrics measuring classification and pose accuracy, it is shown that the presented architecture has desirable robustness and scalable properties. Therefore, the proposed technique can be used to augment the current state-of-the-art monocular vision-based pose estimation techniques used in spaceborne applications.
\end{abstract}

\tableofcontents

\section{Introduction}


The on-board determination of the pose, i.e., the relative position and attitude, of a noncooperative target spacecraft using a monocular camera is a key enabling technology for future on-orbiting servicing and debris removal missions such as e.Deorbit and PROBA-3 by ESA \cite{Castellani2013}, ANGELS by US Air Force \cite{AustrailianDrugFoundation2013}, PRISMA by OHB Sweden \cite{DAmico2012}, OAAN \cite{Pei2017} and Restore-L by NASA \cite{Reed2016}, and CPOD by Tyvak \cite{Bowen2015}. The knowledge of the current pose of the target spacecraft during proximity operations enables real-time approach trajectory generation and control updates \cite{Ventura2016a}. This aspect is crucial in noncooperative maneuvers, since little knowledge about the kinematic characteristics of the target is available before the mission and, therefore, the rendezvous and docking trajectory must be generated on-board using the current state estimates. In contrast to systems based on LiDAR and stereo camera sensors, monocular navigation ensures pose determination under low power and mass requirements \cite{Ventura2015}, making it a natural sensor candidate for navigation systems in future formation flying missions.

The current state-of-the-art monocular pose determination methods for spaceborne applications depend on classical image processing algorithms that identify visible target features \cite{Cropp2002,Benn,Kanani2012,Sharma2015a,Petit2012,DAmico2013,SharmaJSR2017}. These hand-engineered features (e.g., edges, corners, and lines) are then matched against a reference texture model of the spacecraft to determine the pose. This routine is executed in closed-loop for pose tracking using filtering techniques. Generally, the pose solver is an iterative algorithm that minimizes a certain fit error between the features detected in the image and the corresponding features of a reference model. The main advantage of such methods is a high level of interpretation at each step of the image processing and pose determination pipeline. However, these methods are disadvantaged due to the lack of robustness in the presence of adverse illumination conditions and the computational complexity resulting from the evaluation of a large number of possible pose hypotheses. To overcome these two disadvantages, Oumer et al. \cite{Oumer2017} proposed a method based on appearance learning by creating an offline database of feature points and clusters using a vocabulary tree. However, the main drawback of their work is the reliance on a mock-up of the target satellite for training purposes.

In contrast, enabled by availability of large image datasets and cheap computation, the current state-of-the-art pose determination methods for terrestrial applications are shifting towards deep learning techniques \cite{De1997,Khotanzad1996,Thesis2015,Su2016,Wright1992,Sermanet2013,Tompson2014,Toshev2014,Kendall2015}. In particular, recent work \cite{Su2016} proposes to solve a classification problem to determine the pose as opposed to a regression problem. The method is exhibited for a variety of 3D models present in terrestrial environments. This approach can be combined with a sliding-window over an image to solve the detection problem as shown by Romano \cite{Thesis2015} and Sermanet et al. \cite{Sermanet2013}. In comparison to methods used in spaceborne applications, these methods are scalable to tackle multiple types of target in various visual scenes since they do not require the selection of specific hand-engineered features. However, availability of image datasets containing space imagery hinders their use in spaceborne applications. Moreover, unlike imagery captured for terrestrial applications, space imagery is characterized by high contrast, low signal-to-noise-ratio, and low sensor resolution. 

The main contribution of this paper is a deep Convolutional Neural Network (CNN) based pose determination method for spaceborne applications. This work leverages transfer learning and learning based on synthetic space imagery datasets. Additionally, the paper investigates the relationship between the performance of this method and factors such as the size of the training datasets, sensor noise, and the level of pose-space discretization. Finally, the paper also compares the performance of this method against state-of-the-art pose determination methods currently employed for spaceborne applications. The method consists of an off-line training phase and an on-line prediction phase. During the training phase, the method automatically generates several thousand synthetic images of a target spacecraft and uses them to train a CNN. During the prediction phase, the input to the method is a grayscale image of a target satellite taken at close proximity ($\sim$10 [m] inter-satellite separation). The trained CNN is then used to predict a pose label corresponding to a region in the four-dimensional space. Of these four dimensions, three correspond to the attitude of the camera reference frame w.r.t. the target's body reference frame and one corresponds to the distance from the origin of the camera reference frame to the origin of the target's body reference frame. Note that this reduces the problem of estimating the full three-dimensional relative position to a unidimensional relative range. Practically, this implies that the architecture requires a sliding-window based approach \cite{Sermanet2013} to detect the region of the image where the target is present and then use the resulting bearing angle information to re-construct the three-dimensional relative position. Since low-level features (e.g., edges, blobs, etc.) for both terrestrial and spaceborne applications can be hypothesized to be similar, the five convolutional layers of the CNN are trained with images from the ImageNet dataset \cite{Russakovsky2015} while the fully connected layers of the network are trained with synthetically generated images of the Tango satellite of the PRISMA mission. The architecture of the AlexNet network is adopted as the baseline architecture \cite{Krizhevsky2012} for this work.

The paper is organized as follows: Section 2 describes the framework for the synthetic dataset generation and the CNN architecture; Section 3 describes the various combinations of datasets used for training, validation experiments, and accompanying results; and Section 4 presents conclusions from this study and presents directions for further work and development.

\section{Methods}
Formally, the problem statement for this work is the determination of the attitude and position of the camera frame, $C$, with respect to the body frame of the target spacecraft, $B$. In particular, $\mathbf{t_{BC}}$ is the relative position of the origin of the target's body reference frame w.r.t. the origin of the camera's reference frame. Similarly, $\mathbf{q(R_{BC})}$ is the quaternion associated with the rotation matrix that aligns the target's body reference frame with the camera's reference frame. 
\begin{figure}[h]
\includegraphics[width=2.5in,trim={0cm 10cm 10cm 0cm},clip]{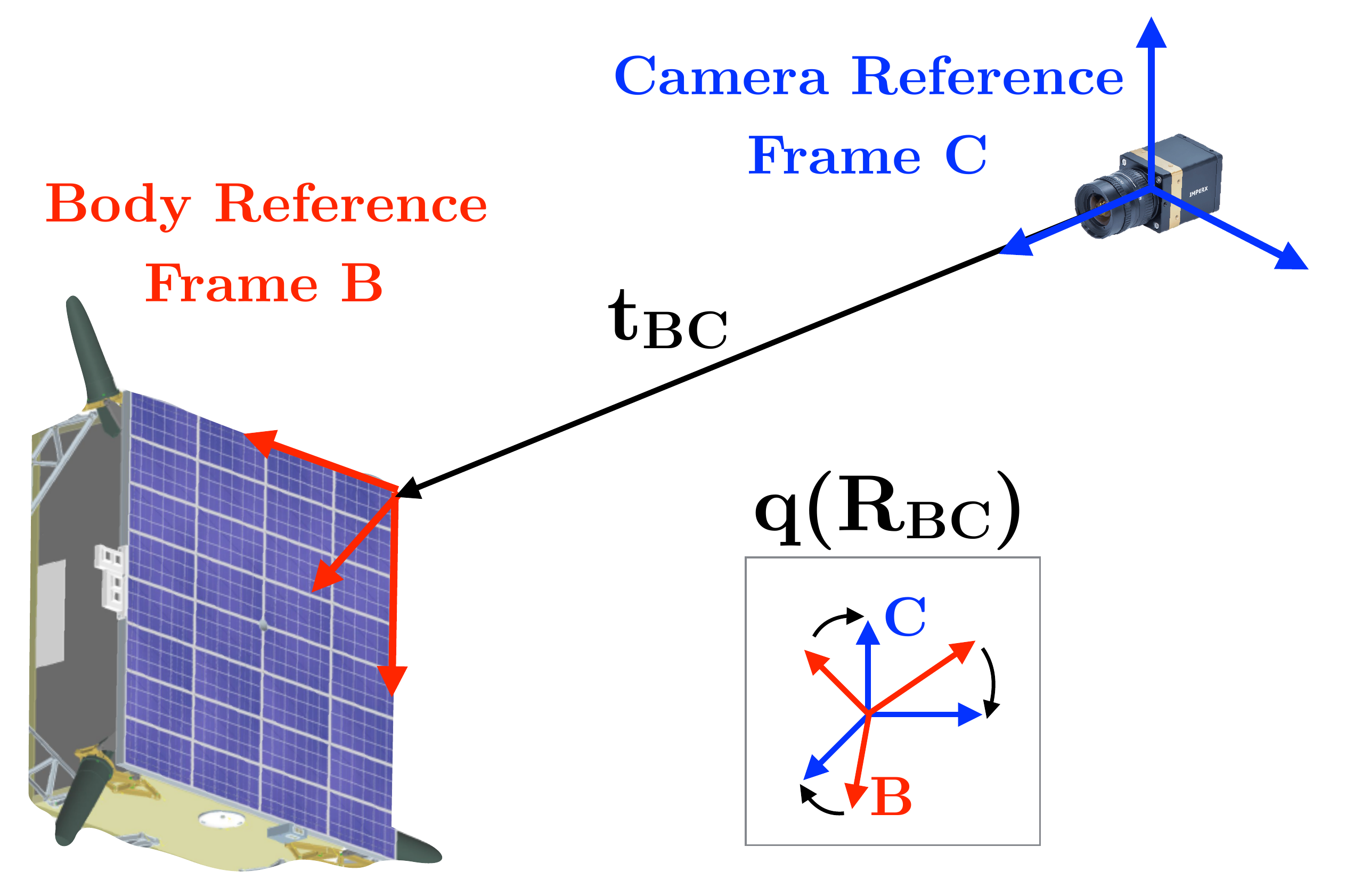}
\centering
\caption{Illustration of the pose determination problem.}
\label{fig:posedet}
\end{figure}

Training a CNN usually requires extremely large labeled image datasets such as ImageNet \cite{Russakovsky2015} and Places \cite{Zhou2014}, which contain millions of images. Collecting and labeling such amount of actual space imagery is extremely difficult. Therefore, this work employs two techniques to overcome this limitation:
\begin{itemize}
	\item a pipeline for automated generation and labeling of synthetic space imagery.
	\item transfer learning which pre-trains the CNN on the large ImageNet dataset.
\end{itemize}
These two techniques are discussed in detail in the following subsections. 

\subsection{Synthetic Dataset Creation}
The automated pipeline for generation and labeling of space imagery is based on discretizing the four-dimensional view-space around a target spacecraft. Three degrees of freedom result from the attitude of the target spacecraft relative to the camera and one degree of freedom results from the distance of the camera from the target spacecraft. 

Uniformly locating a set of $n$ camera locations around the target spacecraft is akin to solving for a minimum-energy configuration for charged particles on a sphere of radius $r$. The determination of a stable configuration of particles constrained on a sphere and being acted by an inverse square repelling force is known as the Thomson problem \cite{Bowick2002}. The solution is a set of $n(n-1)/2$ separations $s_{i,j}$ that minimizes
\begin{equation}
	E = \sum_{i=1}^{n-1}\sum_{j=1+1}^{n}\frac{1}{s_{i,j}}.
\end{equation}
Effectively, a locally optimal solution to the problem can be found by iteratively updating the particle positions along the negative gradient of $E$. A small mesh of camera locations generated in such a manner can be successively subdivided until $n$ camera locations are present on the sphere. Camera locations thus obtained account for two of the four degrees of freedom in the view-space. The third degree of freedom is the rotation of the camera about the boresight direction, which can be uniformly discretized in $m - 1$ intervals from zero to 360$^\circ$. Finally, the degree of freedom corresponding to the distance of the camera relative to the target can be simulated by generating spheres of varying radii. Hence, the inputs to the pipeline are:
\begin{itemize}
	\item Sphere radii, $|\mathbf{t_{BC}}|$
	\item Number of camera locations per sphere, $n$
	\item Number of rotations about the camera boresight per camera location, $m$
	\item 3D texture model of the target spacecraft along with the reflective properties of each of its surfaces and a coarse knowledge of the location of the illumination sources
\end{itemize}
\begin{figure}[h]
\includegraphics[width=\linewidth,trim={19cm 7.1cm 14cm 4cm},clip]{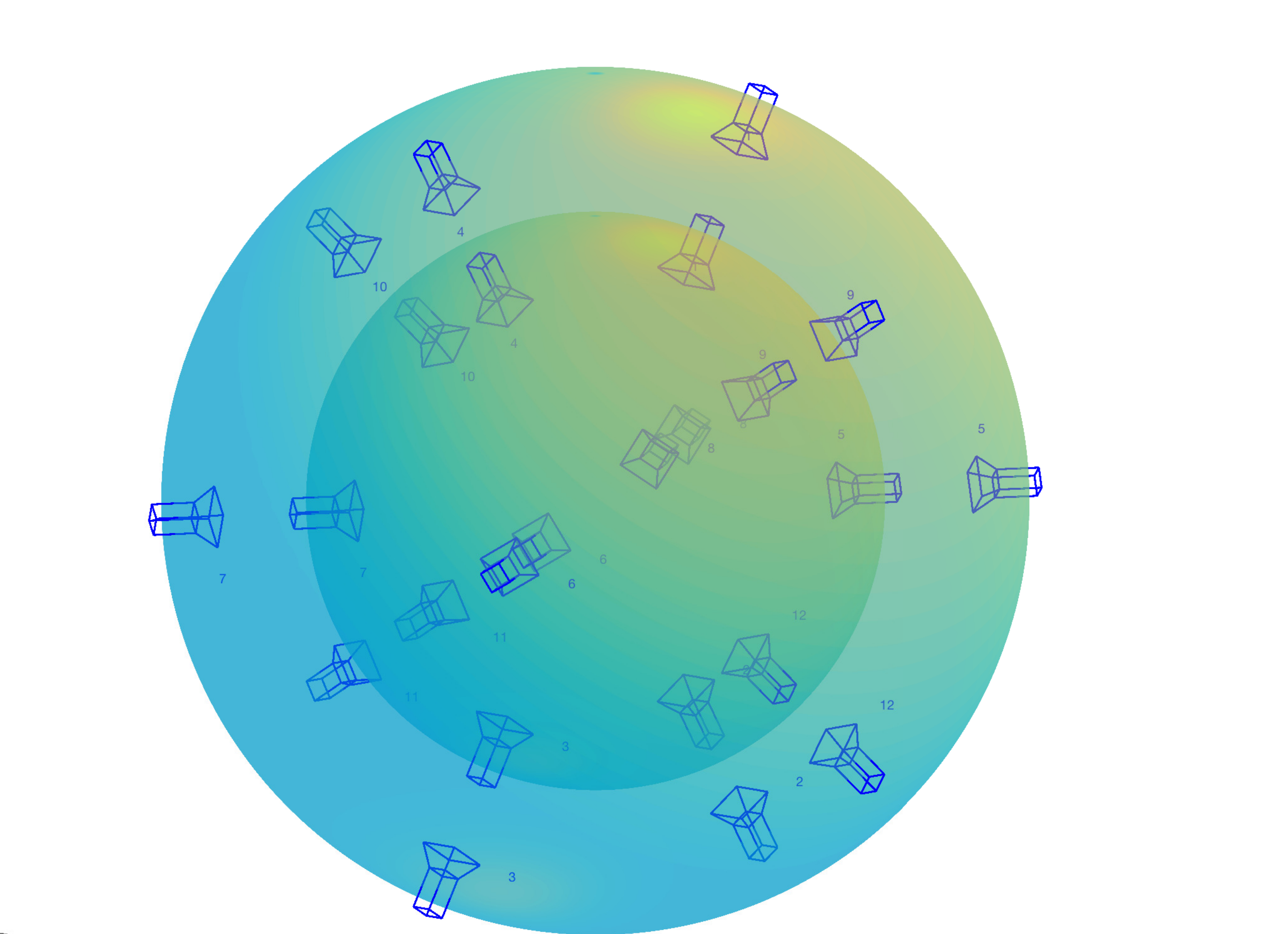}
\centering
\caption{Illustration of the pose space discretization using multiple spheres with uniformly distributed camera locations. This scenario shows two spheres with ten camera locations each.}
\label{fig:posedisc}
\end{figure}

\begin{figure*}
\centering
\includegraphics[width=6in,trim={0 19cm 8cm 0},clip]{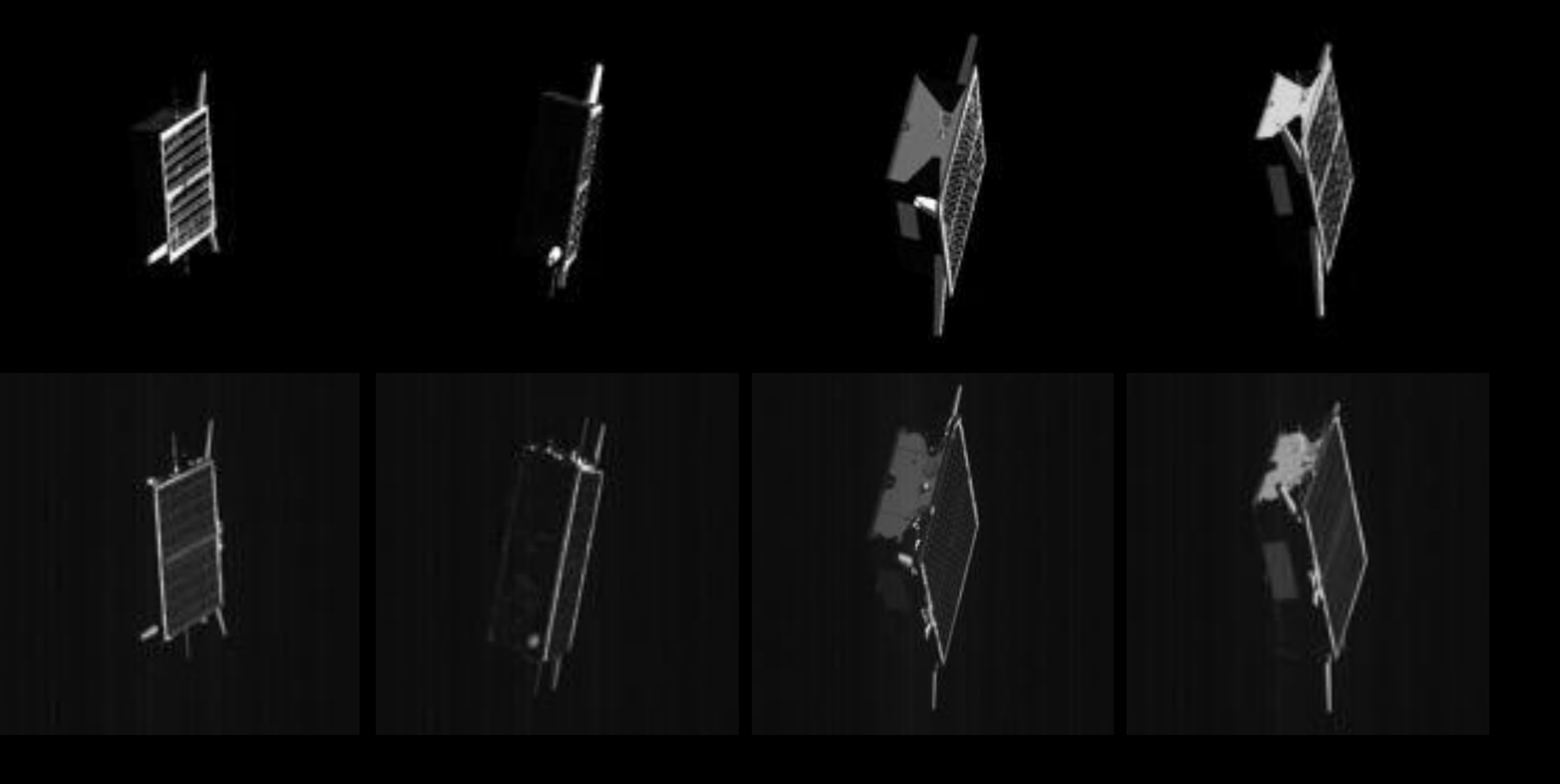}
\caption{Comparison of the synthetically generated images from the Imitation-25 dataset (top row) with actual space imagery (bottom row) from the PRISMA mission. Relative position and orientation of the camera used for image generation were obtained from actual flight data for this dataset.}
\label{fig:realImageComp}
\end{figure*}

Figure~\ref{fig:posedisc} shows a mock scenario with $|\mathbf{t_{BC}}| = 2$, $n = 10$, $m = 1$. To create a total of 125,000 images for the purpose of this paper, the following values were chosen as inputs for the pipeline: $|\mathbf{t_{BC}}| = [8,9,10,11,12,13]$ meters, $n= 500$, $m = 50$. For each of these images, three additional copies were produced with varying levels of Zero Mean White Gaussian Noise (ZMWGN). In particular, the variance of the three levels of noise was selected as 0.01, 0.05, and 0.1 (note that image pixel intensity varies from 0 to 1). Typical images taken in spaceborne applications suffer from high levels of noise due to small sensor sizes and high dynamic range imaging. Therefore, it is imperative to create synthetic images that also possess similar noise characteristics. For each of the 125,000 noise-free images, three additional copies were created in which the target satellite was not aligned with the center of the image plane. This simulates cases where the target spacecraft is in one corner of the image plane, possibly with a few of its features outside the viewing cone of the camera. Finally, a dataset of 25 images (referred to as ``Imitation-25'' in Table \ref{tab:datasets}) was rendered to imitate 25 actual images of the Tango spacecraft from the PRISMA mission \cite{Bodin2012}. Specifically, flight dynamics products from the PRISMA mission consisting of on-ground precise relative orbit determination based on GPS (accurate to about 2 [cm] 3D rms) \cite{Ardaens2012} is used as the relative position. On-board attitude estimates from the Tango spacecraft (accurate to about 3$^\circ$ 3D rms) and the Mango spacecraft (accurate to about 0.1$^\circ$ 3D rms) \cite{DAmico2013} are used to obtain the relative attitude. Note that the Tango spacecraft employed sun sensors and magnetometers while the Mango spacecraft employed a star tracker. Figure \ref{fig:realImageComp} shows a montage of the synthetically generated images part of this dataset compared against their real counterparts.

In total, a superset of 500,000 images were created. The images were rendered using C++ language bindings of OpenGL. Although the pipeline was used to generate synthetic images of the Tango spacecraft used in the PRISMA mission \cite{DErrico2013}, it can easily accommodate any other spacecraft. The camera field of view was selected to be 31.5 degrees, modeling after the close range camera flown aboard the Mango spacecraft of the PRISMA mission. The generated images were resized to be 227 pixels by 227 pixels to match the input size of the AlexNet architecture \cite{Krizhevsky2012} as well as to conserve disk space and RAM usage during the training process. 

After the generation of images, each image must be assigned a pose label that best approximates the true pose of the camera relative to the target spacecraft while capturing the image. This approximate pose label will be used to train the CNN with the expectation that the CNN will learn the visual features associated with the cluster of images belonging to each pose label. More importantly, it is expected that the CNN will learn the correlation between these learned features and the approximate pose label associated with those images. Since the CNN solves a classification problem to determine a pose that exists in a continuous domain, it is important to clarify the distinction between classification and pose estimation accuracies. It is expected that the level of discretization of the pose space used during training drives the accuracy of the on-line pose estimation. Thus, the choice of the number of pose labels used during training depends on the required pose estimation accuracy. For the purpose of this paper, four levels of discretization were used resulting in 6, 18, 648, and 3000 pose labels. The pose labels were generated using the same procedure as described above for the image generation. The input values $|\mathbf{t_{BC}}|$, $n$, $m$ used for each of these pose labels is presented in Table~\ref{tab:poselabels}.
\begin{table}
\renewcommand{\arraystretch}{1.3}
\caption{\bf Summary of discretization levels used to generate the pose labels.}
\label{tab:poselabels}
\centering
\begin{tabular}{|c|c|c|c|}
\hline
\bfseries \# Pose Labels & $\mathbf{|\mathbf{t_{BC}}|}$ [m] & \bfseries $\mathbf{n}$ & \bfseries $\mathbf{m}$\\
\hline\hline
6	& \{3\}		& 6	 & 1 \\
18	& \{3, 5, 9\}	& 6 & 1 \\
648 & \{8,9,10,11\} & 162 & 1 \\
3000 & \{8,9,10,11,12\} & 300 & 10\\
\hline
\end{tabular}
\end{table}
Each generated image is then assigned to a pose label for each of the four levels of discretization using a simple search algorithm. First, all pose labels associated with the same camera distance relative to the target as the image are selected. Then, for each possible pose label an axis-angle parametrization of the attitude change required to match the camera attitude associated with the image can be calculated. Finally, the pose label that minimizes this angular change is selected as that image's pose label. The pseudo-code for this search algorithm is presented below as Algorithm \ref{alg:poselabel}. 
\begin{algorithm}
\caption{Assigns a pose label to an image for each of the four levels of pose discretization}\label{alg:poselabel}
\begin{algorithmic}[1]
\Procedure{assignLabel}{image,allPoseLabels}
\State image.label = NaN
\State minAngDiff = Inf
\For{label in allPoseLabels}
	\If{image.dist == label.dist}
    	\State quatDiff = \texttt{quatmult}(image.quat, quatinv(label.quat))
    	\State angDiff = \texttt{quat2axang}(quatDiff)
        
        \If{angDiff $<$ minAngDiff}
        	\State minAngDiff = angDiff
            \State image.label = label
    	\EndIf    
    \EndIf	
\EndFor
\EndProcedure
\end{algorithmic}
\end{algorithm}
\begin{figure}[h!]
\includegraphics[width=\linewidth,trim={2cm 0cm 2cm 0cm},clip]{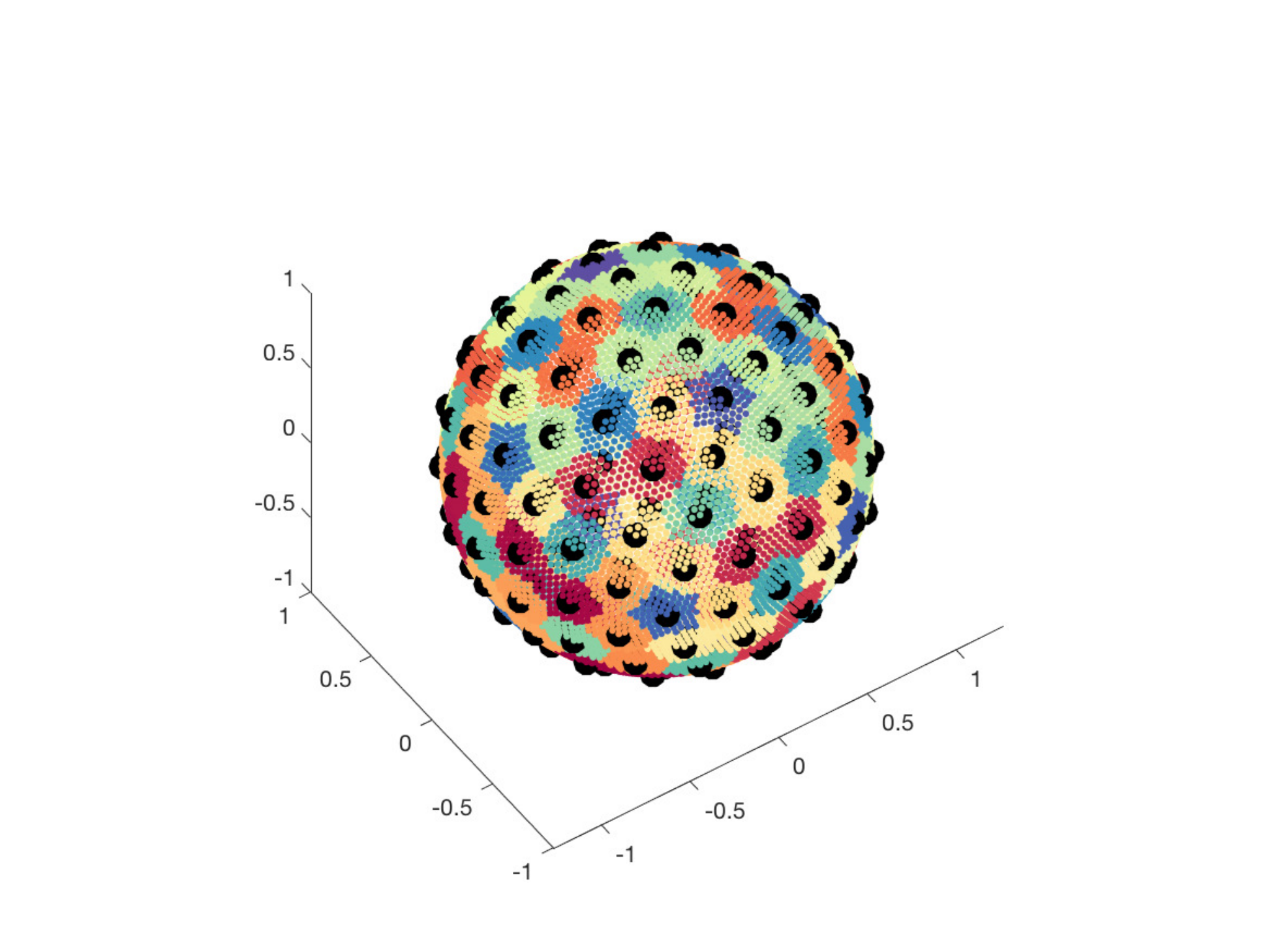}
\centering
\caption{Visualization of a uniform distribution of 10242 camera locations (small colored markers) and 162 pose labels (large black markers) around a unit sphere. Camera locations associated to the same pose label are denoted by the same colored marker.}
\label{fig:648ClassDisc}
\end{figure}

\begin{figure*}
\centering
  \mbox{\subfloat[]{\label{subfig:a} \includegraphics[trim={0 10cm 40cm 0},clip,width=3in]{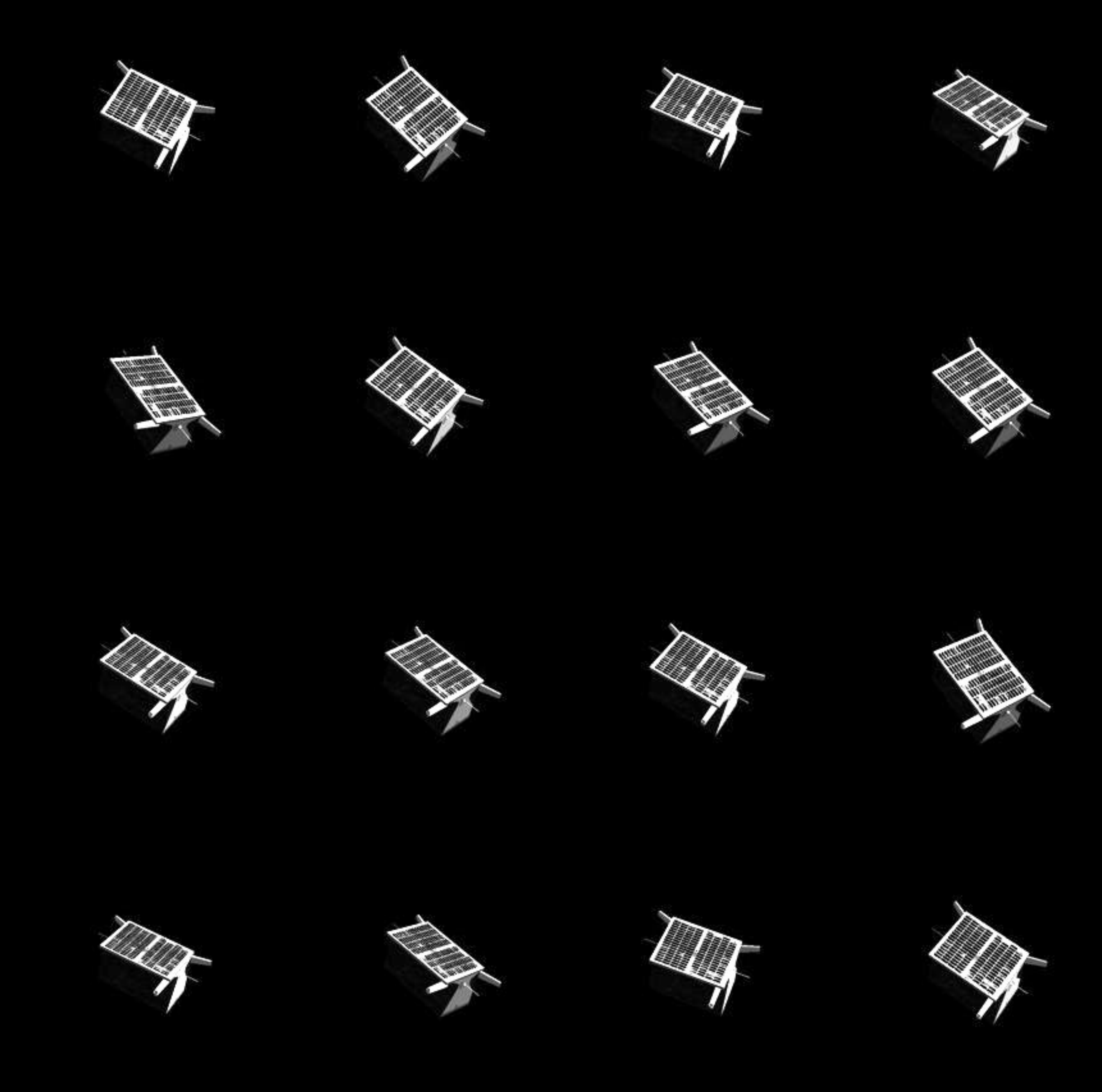}}}
  \mbox{\subfloat[]{\label{subfig:b} \includegraphics[trim={0 10cm 40cm 0},clip,width=3in]{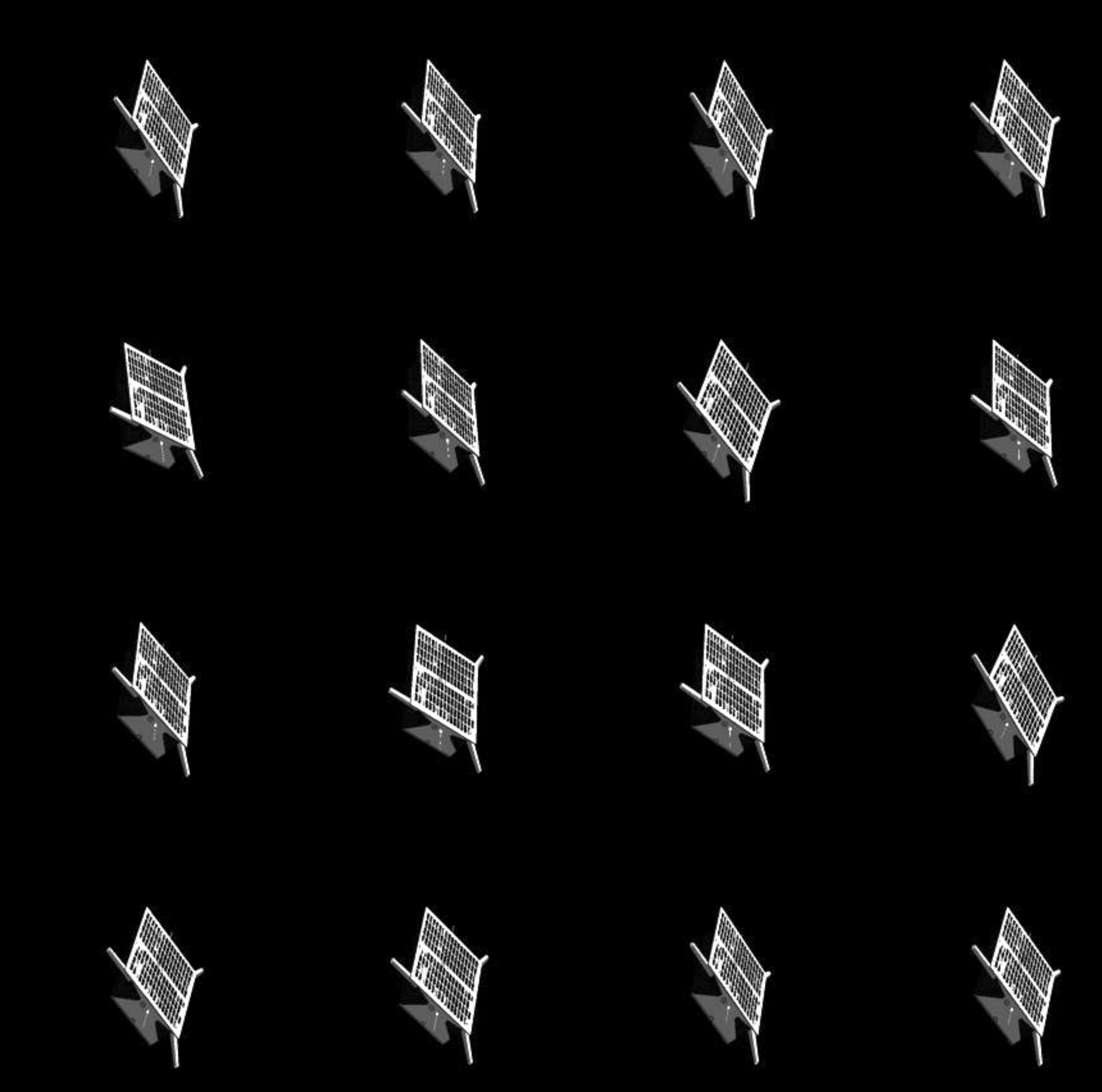}}}
  \mbox{\subfloat[]{\label{subfig:b} \includegraphics[trim={0 10cm 40cm 0},clip,width=3in]{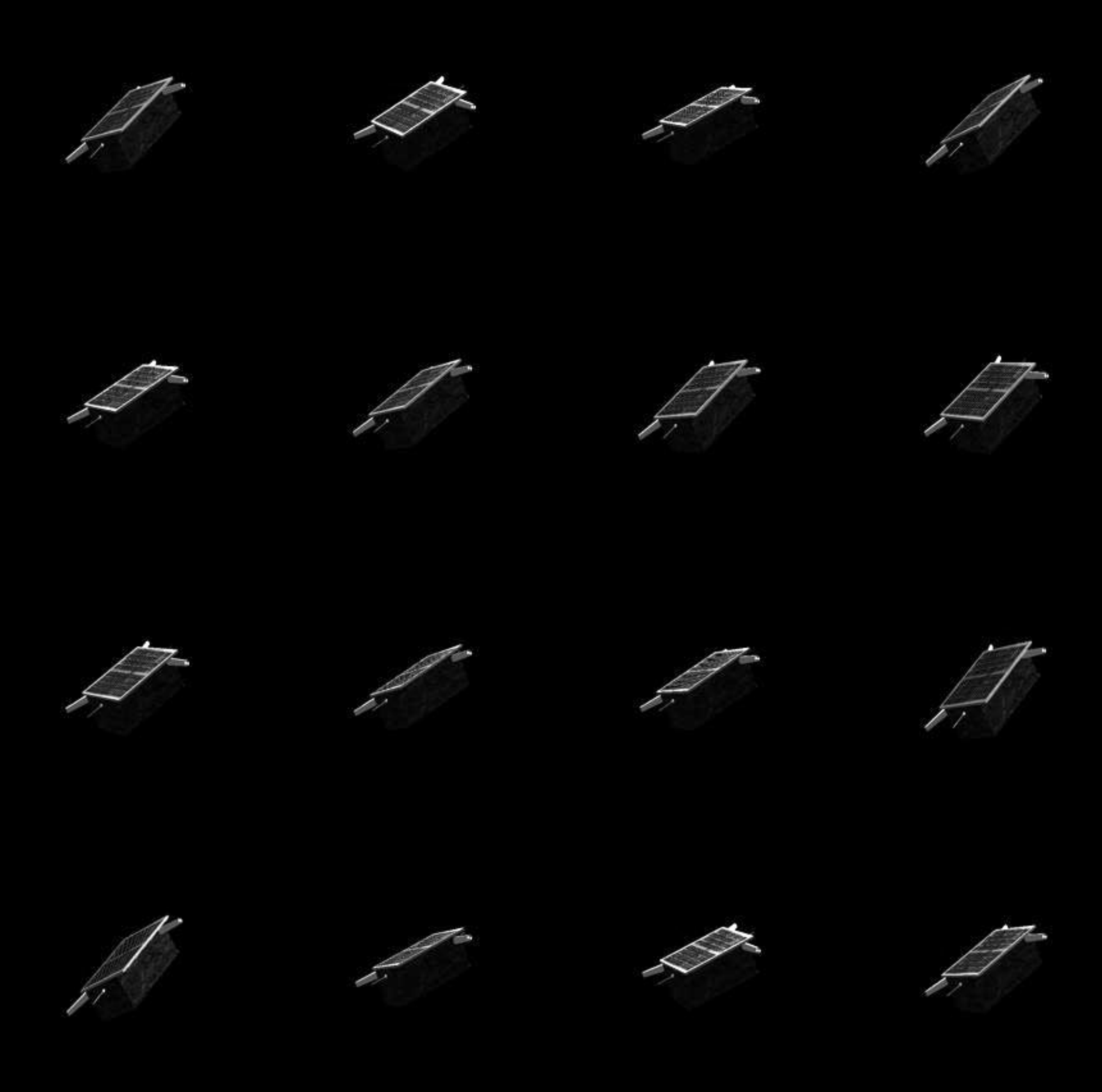}}}
  \mbox{\subfloat[]{\label{subfig:b} \includegraphics[trim={0 10cm 40cm 0},clip,width=3in]{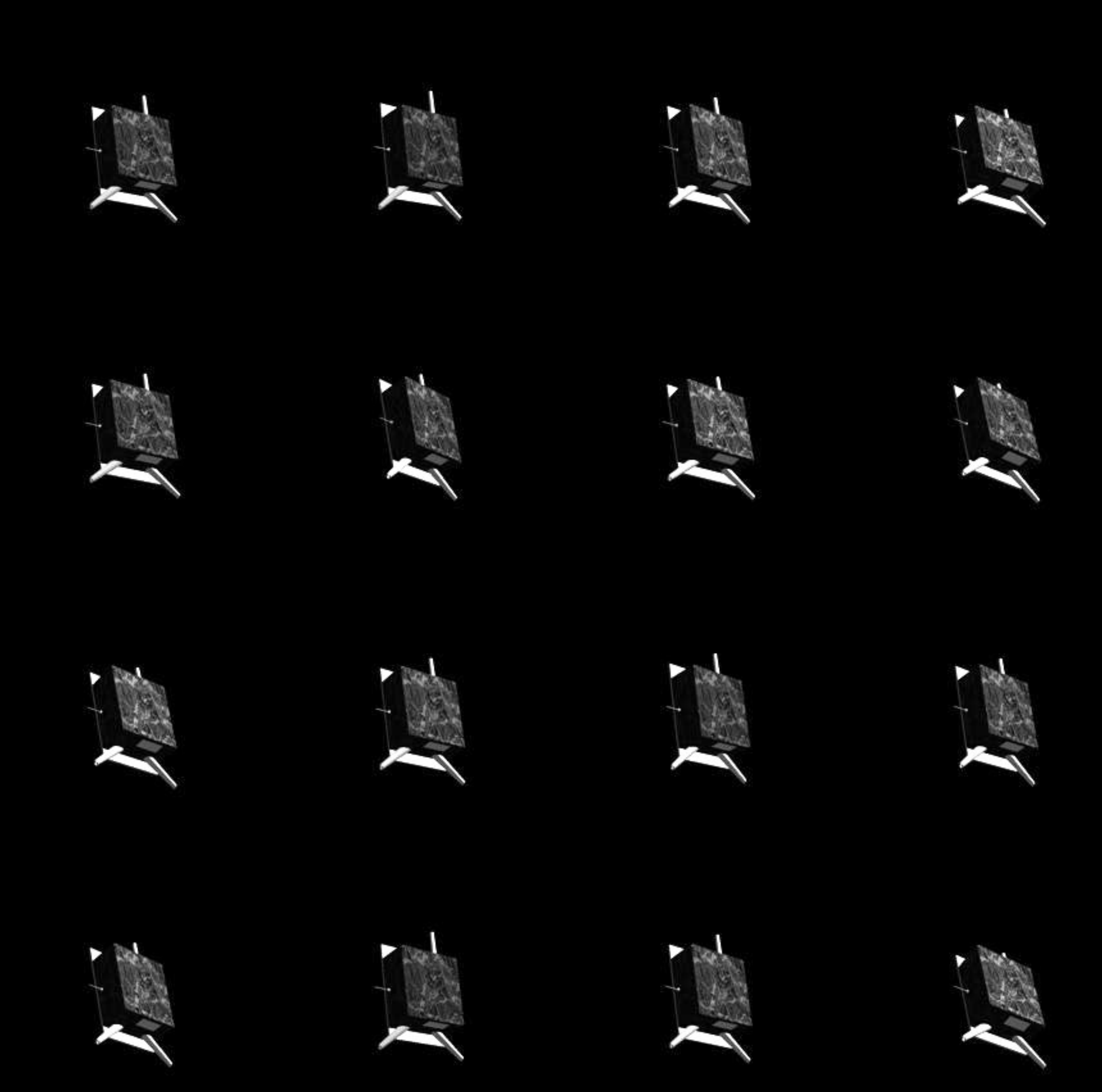}}}
  \caption{Montage of a few images from four different pose labels of the Clean-648 dataset.}
  \label{fig:montage}
\end{figure*}

Note that Algorithm \ref{alg:poselabel} is repeated for each level of discretization. For the purpose of this paper, this allowed us to compose the superset of 500,000 images into 10 datasets. Table \ref{tab:datasets} presents the details for each of these datasets. Each dataset was further divided into a training, validation, and test set, which represented 60\%, 20\%, and 20\% images of the dataset, respectively. Figure \ref{fig:montage} shows a montage of images associated with four different pose labels of the Clean-648 training dataset.

\begin{table*}
\renewcommand{\arraystretch}{1.3}
\caption{\bf Description of the ten datasets created from the synthesized images.}
\label{tab:datasets}
\centering
\begin{tabular}{|c|c|c|c|}
\hline
\bfseries Dataset& \bfseries Description & \bfseries \# Pose Labels & \bfseries \# Images\\
\hline\hline
Clean & No noise, centered target & 6 & 1601\\
Clean-18 & No noise, centered target & 18 & 4803\\
Clean-648 & No noise, centered target & 648 & 40968\\
Clean-3k & No noise, centered target & 3000 & 125000\\
Gaussian-1 & ZMGWN with variance of 0.01, centered target & 6 & 1601\\
Gaussian-5 & ZMGWN with variance of 0.05, centered target & 6 & 1601\\
Gaussian-10 & ZMGWN with variance of 0.1, centered target & 6 & 1601\\
Centered-1 & No noise, $t= [0.2, 0.0, 3.0]$ meters & 6 & 1601\\
Centered-2 & No noise, $t= [0.0, 0.2, 3.0]$ meters & 6 & 1601\\
Centered-3 & No noise, $t= [0.2, 0.2, 3.0]$ meters & 6 & 1601\\
Imitation-25 & No noise, PRISMA flight data & 3000 & 25\\
\hline
\end{tabular}
\end{table*}

\subsection{Convolutional Neural Network}
The CNN used in this work adopts the structure of the AlexNet architecture~\cite{Krizhevsky2012}. AlexNet was chosen over networks such as VGG-16 \cite{Simonyan2014} and Inception \cite{Szegedy2016} due to the relatively lower number of operations required for inference \cite{Canziani2016}. Moreover, since the number of images in the synthetically generated datasets is not as high as typical datasets used to train these networks, the proposed method relies on transfer learning. The hypothesis is that low level features detected by the first few layers of a CNN are the same across the terrestrial and spaceborne domains. Therefore, only the parameters in the last few layers need to be determined to adapt the network for space imagery. The AlexNet architecture was used to train eight networks with varying sizes and compositions of the training set. The description of the eight networks is presented in Table \ref{tab:nets}.

\begin{table*}[h!]
	\renewcommand{\arraystretch}{1.3}
    \caption{\bf{Description of the eight networks trained for this work. Note that the columns represent the number of training images used from the particular dataset.}}
    \label{tab:nets}
    \centering
    \begin{tabular}{|c|c|c|c|c|c|c|}
    	\hline
    	\bfseries Network & \bfseries \# Pose Labels & \bfseries Clean & \bfseries Gaussian-1 & \bfseries Clean-18 & \bfseries Clean-648 & \bfseries Clean-3k \\
    	\hline \hline 
        net1         & 6		& 873     & 0            & 0		& 0			& 0 \\
		net2         & 6		& 728     & 0            & 0		& 0			& 0 \\
		net3         & 6		& 582     & 0            & 0		& 0			& 0 \\
		net4         & 6		& 436     & 0            & 0		& 0			& 0 \\
		net5         & 6		& 873     & 436          & 0		& 0			& 0 \\
		net6         & 18		& 0       & 0            & 2619		& 0			& 0 \\
		net7         & 648		& 0       & 0            & 0		& 24581			& 0 \\
		net8         & 3000		& 0       & 0            & 0		& 0			& 75000 \\
        \hline
    \end{tabular}
\end{table*}

\begin{figure*}[h]
\includegraphics[width=0.7\linewidth]{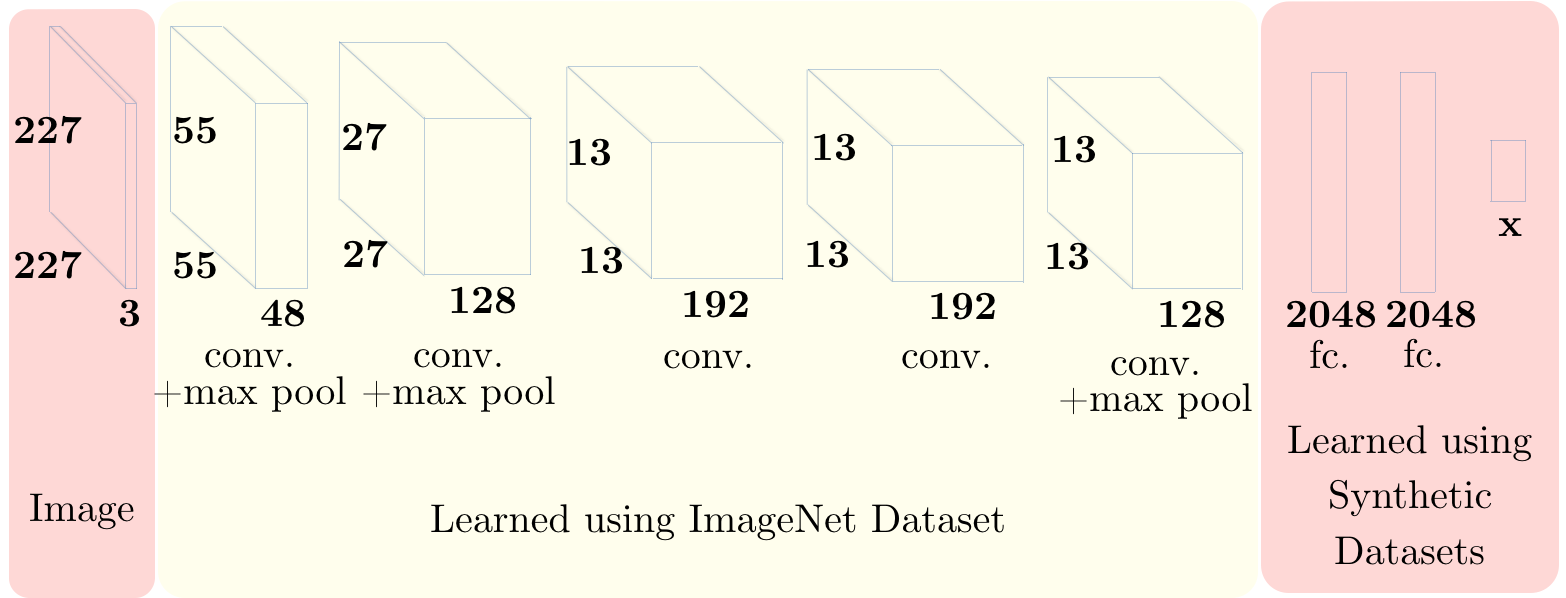}
\centering
\caption{An illustration of the architecture of AlexNet, used as the baseline for all eight networks in this paper. The network's input is 154587-dimensional, and the number of neurons in the network's remaining layers is given by 145200--93312--32448--32448--21632--2048--2048--x. The last layer contains as many neurons as the number of pose labels in the dataset used to train the particular network.}
\label{fig:alexnet}
\end{figure*}

The AlexNet architecture is shown in Figure \ref{fig:alexnet}, it contains eight layers with weights, the first five are convolutional layers and the remaining three are fully-connected layers. The output of the last fully-connected layer is used in an $x$-way softmax loss function which produces a distribution over the class labels (where $x$ is the number of pose labels in the dataset used to train the network). The formula to compute the softmax loss function is presented below in equations \ref{eq:softmax1} and \ref{eq:softmax2}. The network maximizes the multinomial logistic regression objective, which is equivalent to maximizing the average across training cases of the log-probability of the correct label under the prediction distribution.
\begin{equation}
	\text{Loss}_\text{softmax} = \sum_{i=1}^x L_i 
	\label{eq:softmax1}
\end{equation}
where
\begin{equation}
	L_i = -\log \left ( \frac{e^{f_{y_i}}}{\sum_j e^{f_j}} \right )
	\label{eq:softmax2}
\end{equation}
In equation \ref{eq:softmax2} the notation $f_j$ refers to the j-th element of the vector of values output by the last fully-connected layer, $f$. 

The ``dropout'' technique \cite{Hinton2012} was used while training the fully connected layers. This technique consists of setting to zero the output of each hidden neuron with probability of 0.5. The ``dropped'' neurons do not contribute to the forward pass and do not participate in back-propagation. The output of the remaining neurons is scaled by a factor of 2 such that the expected sum remains unchanged. This technique reduces the possibility of co-adaptations of neurons, i.e., neurons cannot rely on the presence of particular other neurons but instead learn more robust features. Secondly, horizontal reflection of images was utilized for ``net1'',``net2'',``net3'',``net4'', and ``net5'', which effectively increased the size of the training set by a factor of two. This data augmentation technique was hypothesized to reduce over-fitting on the image data by artificially enlarging the dataset.

\section{Experiments}

This section presents results of two types of experiments. The first type of experiments were carried to understand the training process of the CNN's on small sets of images and evaluating the network's test accuracy as a factor of:
\begin{itemize}
	\item Number of images used in training
	\item Amount of ZMGWN in test images
	\item Amount of displacement of the target from the center of the image plane
\end{itemize}
The second set of experiments were carried to understand the feasibility of training CNN's for a realistic on-orbit servicing mission and evaluating its performance on imagery imitating actual space imagery from the PRISMA mission. Both of these experiments and their results are discussed in the following subsections.

\subsection{Type 1}
The Type 1 experiments involved comparing the performance of net1, net2, net3, net4, net5, and net6 on the Clean-6, Clean-18, Gaussian-1, Gaussian-5, Gaussian-10, Centered-1, Centered-2, and Centered-3 datasets. The comparison is based on the accuracy of the predictions and the F-Measure ($FM$) of the classifications. The results presented here are based on testing the networks on the test set of the datasets described in Table \ref{tab:datasets}. In particular, the accuracy is defined as the percentage of the test images that were correctly classified by the network. $FM$ of the classifications is based on the precision and recall. These metrics are based on the number of false positives ($FP$), false negatives ($FN$), and true positives ($TP$) over several samples. In order to compute these values, we treat each class as a binary classification problem, defining a positive sample when it belongs to that class, and negative otherwise. 
\begin{equation}
	\text{precision} = \frac{TP}{(TP + FP)}
\end{equation}
\begin{equation}
	\text{recall} = \frac{TP}{(TP + FN)}
\end{equation}
These are then used to calculate the F-Measure:
\begin{equation}
	FM = 2 \cdot \frac{\text{precision} \cdot \text{recall}}{\text{precision} + \text{recall}}
\end{equation}

\begin{figure*}[h]
\includegraphics[width=\linewidth]{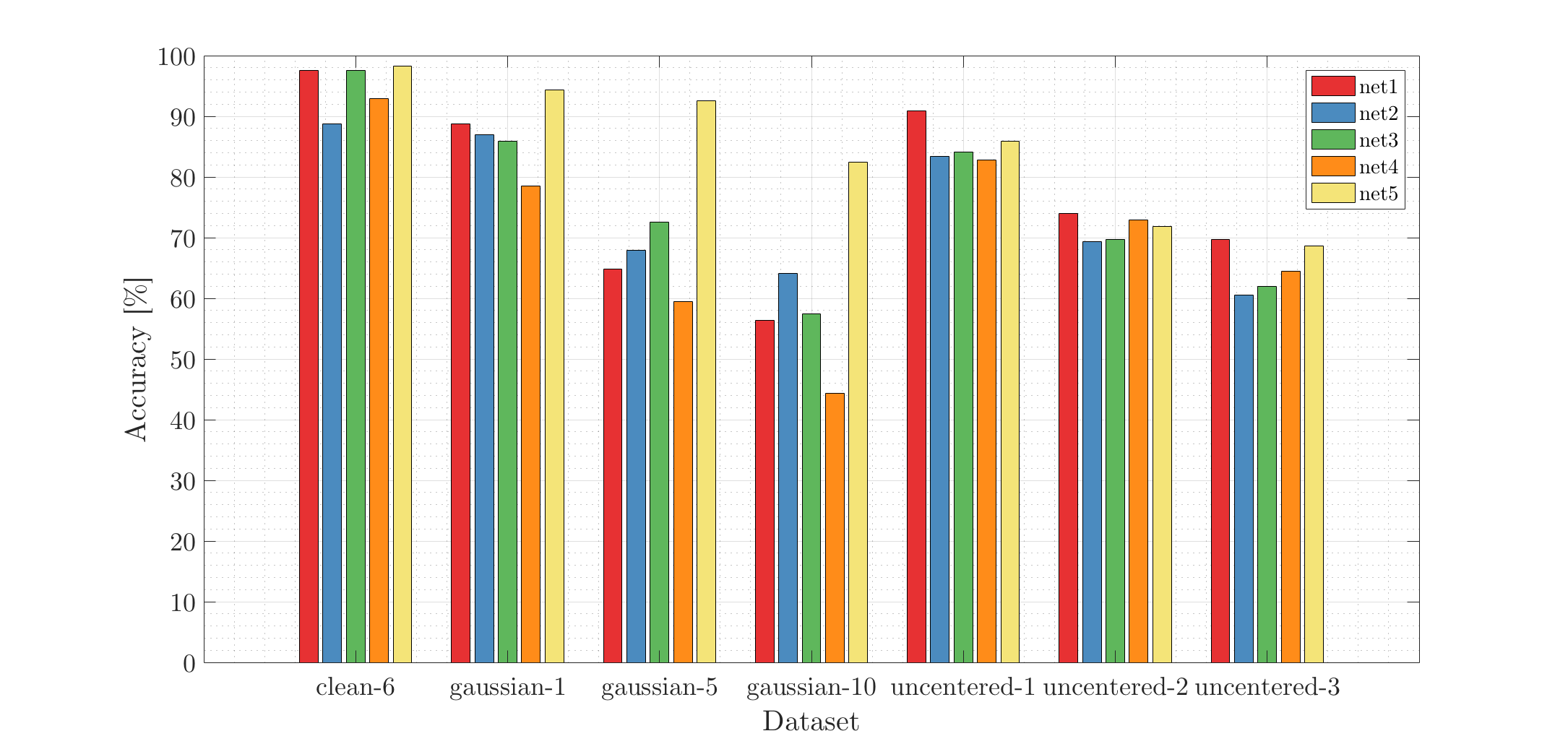}
\centering
\caption{Classification accuracy [\%] for seven datasets using five separate networks.}
\label{fig:accVsDataset}
\end{figure*}

There are several trends as seen in Figure~\ref{fig:accVsDataset}, which shows the classification accuracy of five separately trained networks on six different datasets. Firstly, all networks are more or less equally capable in classifying images in the Clean-6 dataset where images are free from ZMGWN and the target is centered in the image plane (all networks trained in this work were trained with centered targets). This most likely shows that the networks have a vast number of parameters and can easily over-fit the features seen in the images. Secondly, as the ZMGWN is added, all networks show a decline in the classification accuracy. Notably, ``net5'' fares quite well as compared to the other networks as it used some noisy images during training. This implies that as long as sensor noise is known and can be modeled beforehand, the CNN can be made to be more robust to noise through the augmentation of the training data with noise. Thirdly, the classification accuracy of the networks correlates with the number of training images used during the training since ``net1'', which was trained with more images than ``net2'', ``net3'', and ``net4'' has higher accuracy for Clean-6, Uncentered-1, Uncentered-2, and Uncentered-3 datasets.

\begin{figure}[h]
\includegraphics[width=\linewidth,trim={0 0cm 0cm 1cm},clip]{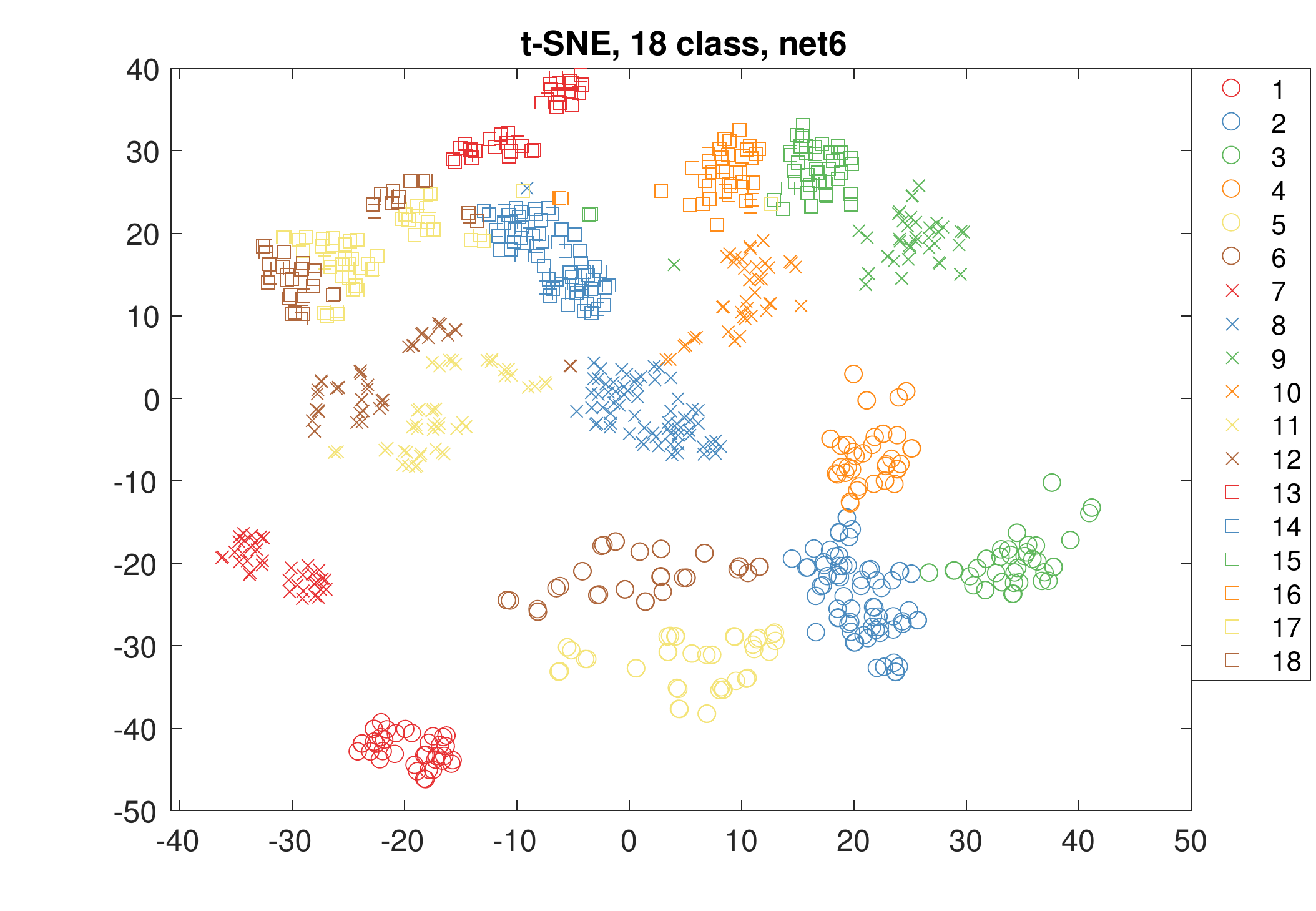}
\centering
\caption{The t-Distributed Stochastic Neighbor Embedding (t-SNE)representation of the Clean-18 test set images.}
\label{fig:tsne}
\end{figure}

Lastly, ``net6'' was trained and tested using the Clean-18 dataset without data augmentation. The network produced a classification accuracy of 99.4\%, which was significantly higher than any other networks trained on smaller datasets with data augmentation. To visualize how the network had learned to separate the 18 different classes, the test set images of Clean-18 were embedded according to their features from the penultimate fully connected layer. The t-Distributed Stochastic Neighbor Embedding (t-SNE) technique was used for dimensionality reduction \cite{maaten2017}. This technique represents images as nodes in a graph and arranges the nodes in two dimensions in a manner that respects the high-dimensional L2 distances between their features. In other words, t-SNE arranges images that have similar features nearby in a 2D embedding. This can be visualized in Figure~\ref{fig:tsne} for the Clean-18 test set images, where images from three different inter-satellite ranges are represented by different marker types. It can be easily seen that the network has learned to differentiate images from separate ranges. Moreover, for each range, certain classes are learned to be closer as compared to the others. This is to be expected since for example, two pose labels visually do look similar (in fact, they are close to being horizontal mirrors of each other).

\subsection{Type 2}
The Type 2 experiments involved evaluating the performance of net6 and net7 on the Clean-648, Clean-3k, and Imitation-25 datasets. The goal of these experiments was to stress-test all key aspects of this method, from conceiving pose labels, to training and testing. Unlike the Type 1 experiments, these experiments were run with a high number of pose labels and large training datasets in order to achieve a higher ``pose estimation accuracy''. In particular, the pose estimation accuracy is defined by two metrics: $E_R$ and $E_T$, which are differences between true and estimated values of relative attitude and position, respectively. In particular,
\begin{eqnarray}
	E_T &=& \left \| \mathbf{t_{BC,\text{est}}} - \mathbf{t_{BC,\text{true}}} \right \| \\
	E_R &=& 2 \cos^{-1}(z_s), \text{where}	\\
	\mathbf{z} &=&  \left [  z_s \; \mathbf{z_v}  \right ]= \mathbf{q}_\text{true} * \text{conj}(\mathbf{q}_\text{est}). \nonumber
\end{eqnarray}
Here $\mathbf{q}_\text{true}$ and $\mathbf{q}_\text{est}$ are true and estimated values of the quaternion associated with the rotation matrix that aligns the target's body reference frame and the camera's reference frame. Table~\ref{tab:net78onSyn} shows the test set accuracy for net7 and net8 on the Clean-648 and Clean-3k datasets, respectively. Note that net7 has a much higher classification accuracy as compared to net8 since it only needs to pick the correct pose label out of a set of 648 pose labels compared to 3000 for net7. In addition, net7 was trained on the Clean-648 dataset which contained approximately 45 images per pose label as compared to 25 images per pose label for the Clean-3k dataset used for net8. However, due to the larger number of classes in Clean-3k dataset, net8 produced higher pose estimation accuracy compared to net7. 
\begin{table}
\renewcommand{\arraystretch}{1.3}
\caption{\bf Performance of the net7 and net8 networks on test sets of the Clean-648 and Clean-3k datasets, respectively.}
\label{tab:net78onSyn}
\centering
\begin{tabular}{|c|c|c|}
\hline
\bfseries Metric & \bfseries net7 & \bfseries net8 \\
\hline\hline
Mean $E_R$ (deg)		& 22.91		& 11.94 \\
Mean $E_T$ (m)	 	& 0.53 & 0.12 \\
Mean Classification Accuracy (\%) & 83.3 & 35 \\
\hline
\end{tabular}
\end{table}

\begin{figure*}
\centering
  \mbox{\subfloat[]{\label{subfig:a} \includegraphics[trim={4cm 0cm 4cm 0cm},clip,width=3in]{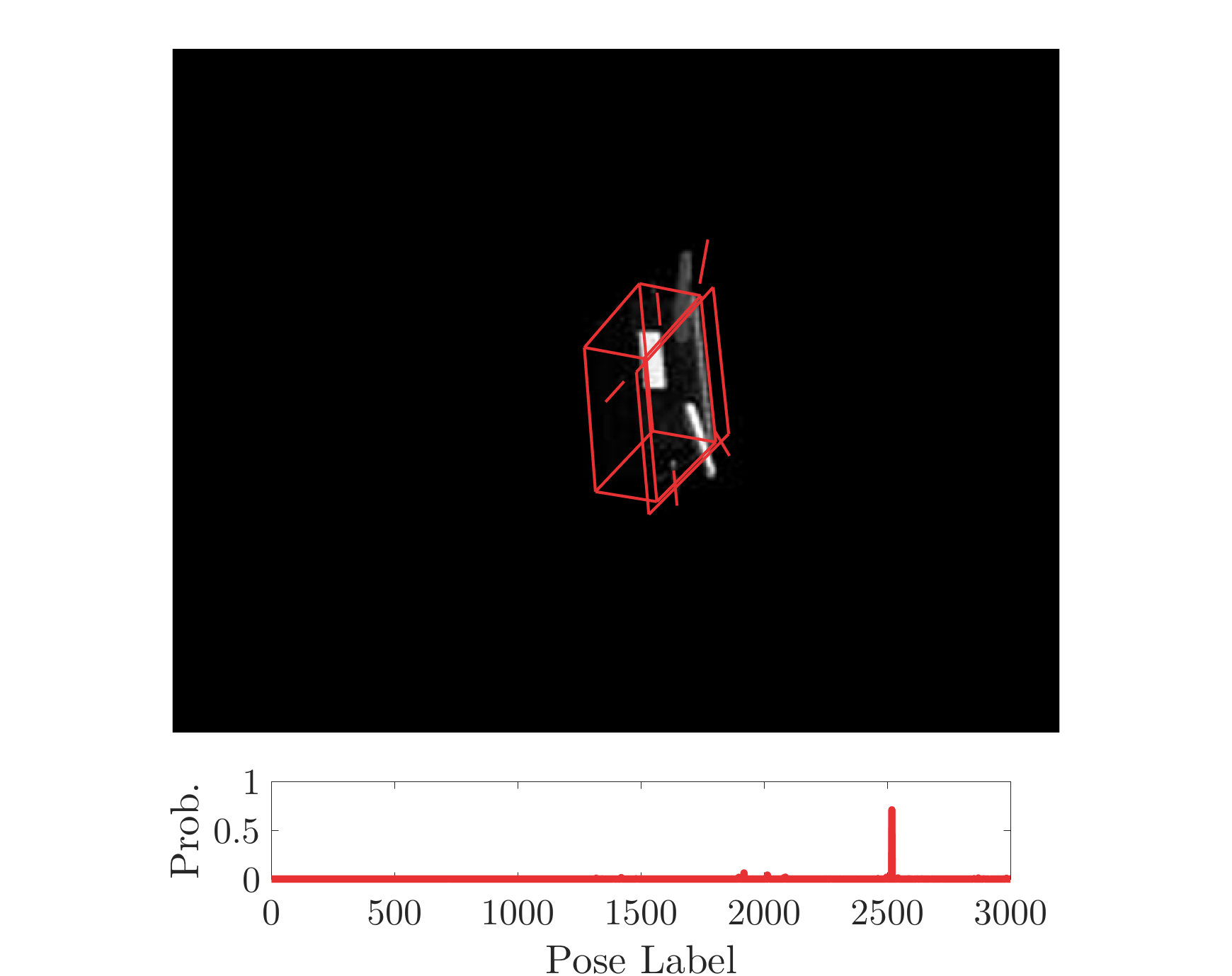}}}
  \mbox{\subfloat[]{\label{subfig:b} \includegraphics[trim={4cm 0cm 4cm 0cm},clip,width=3in]{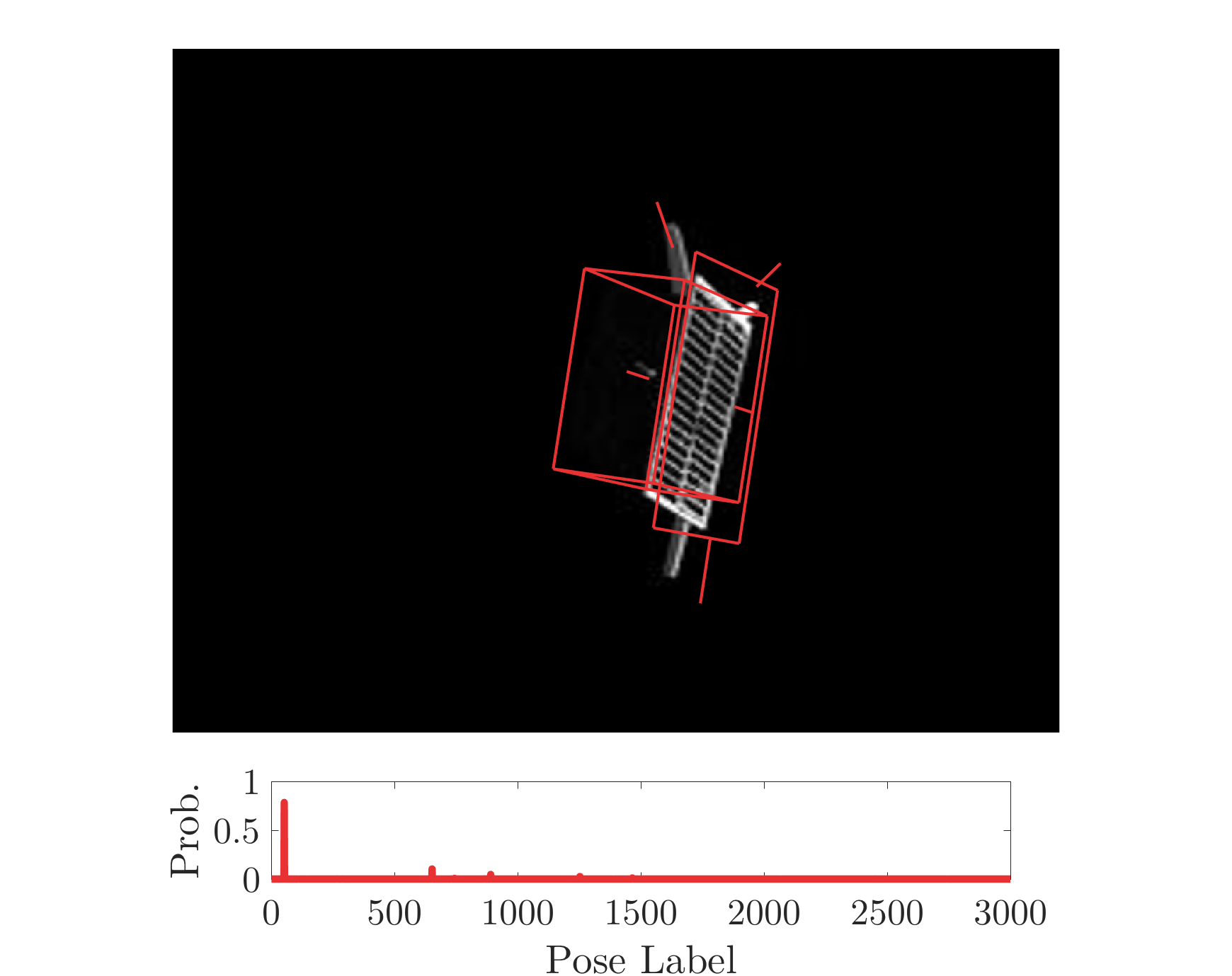}}}
  \mbox{\subfloat[]{\label{subfig:b} \includegraphics[trim={4cm 0cm 4cm 0cm},clip,width=3in]{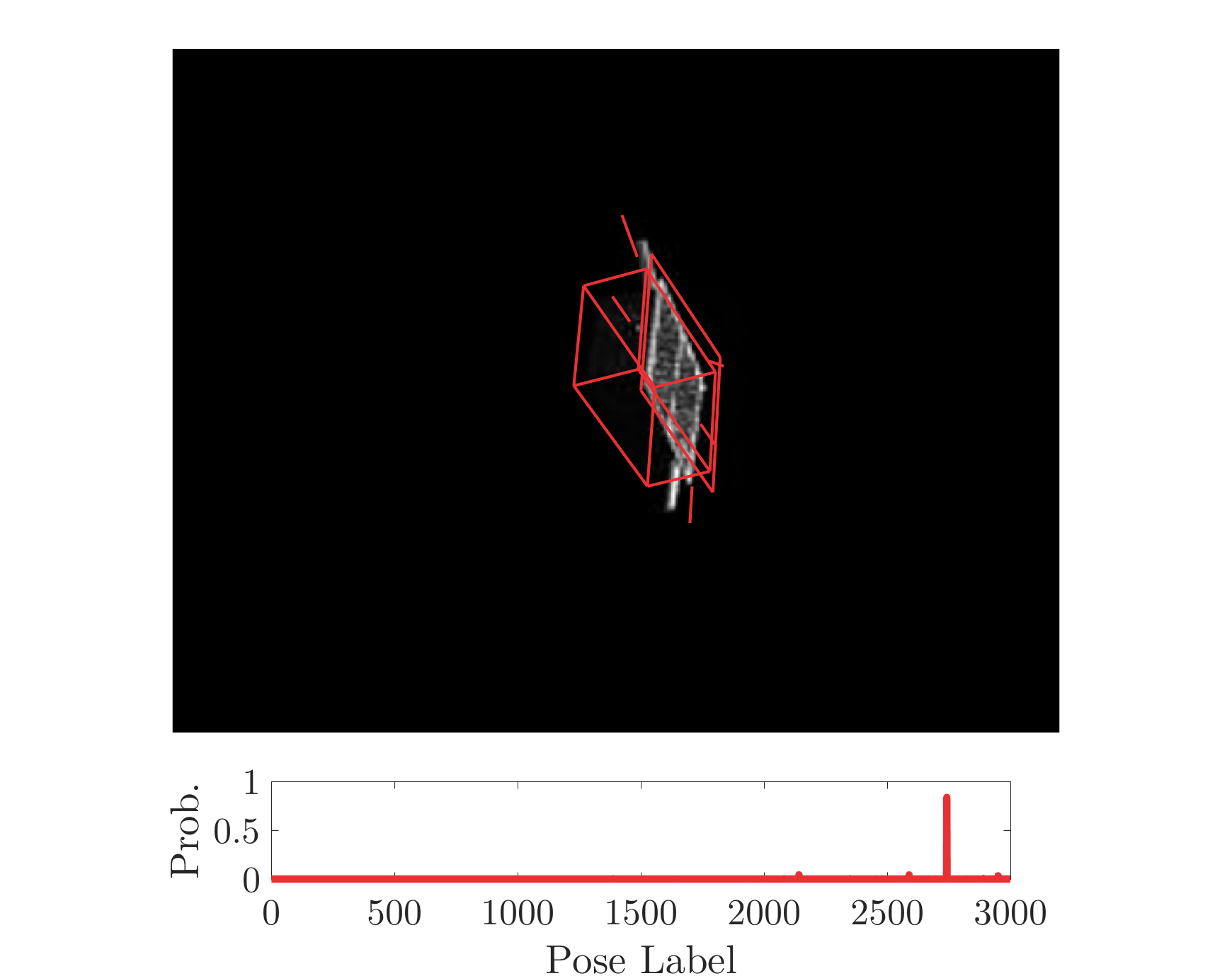}}}
  \mbox{\subfloat[]{\label{subfig:b} \includegraphics[trim={4cm 0cm 4cm 0cm},clip,width=3in]{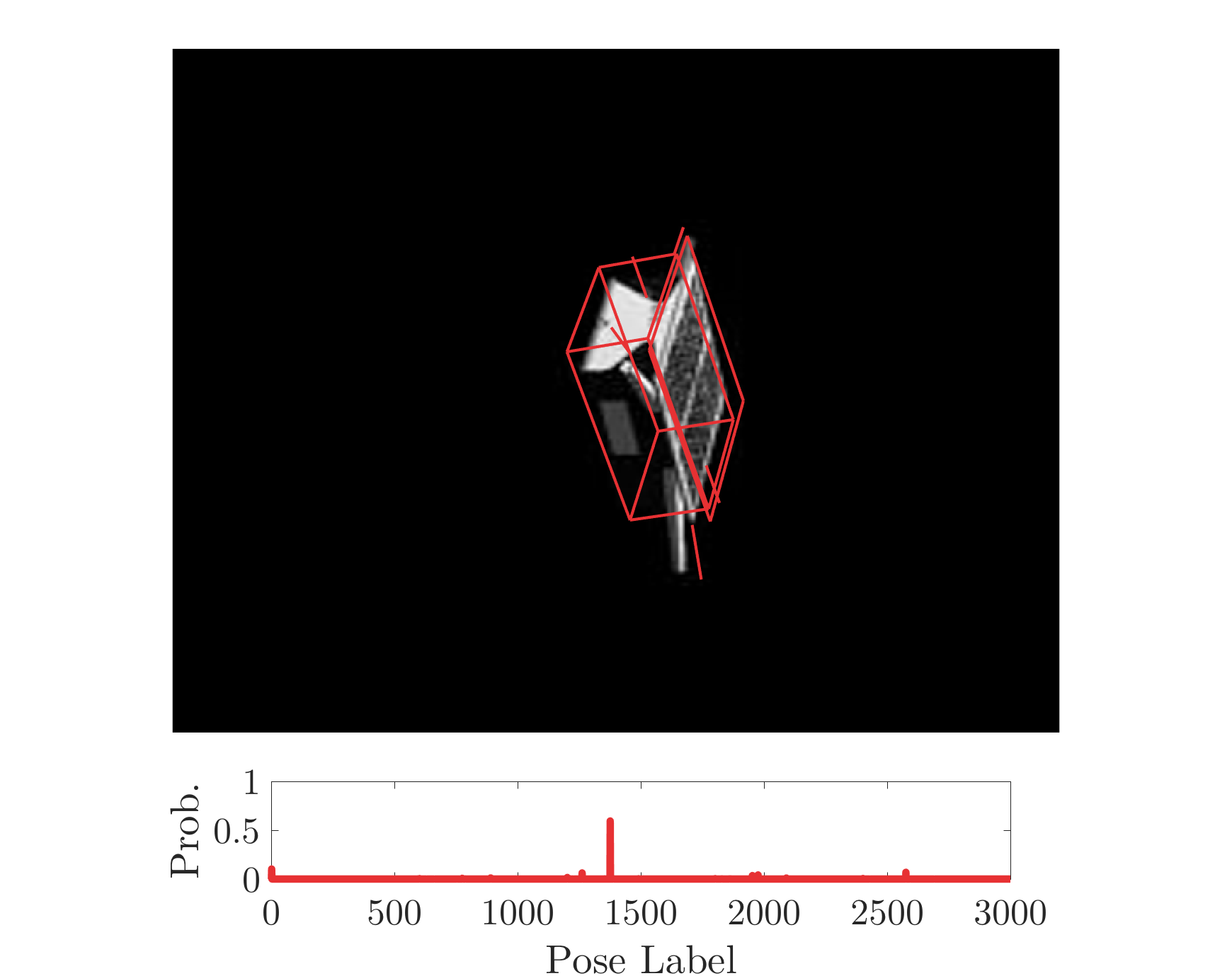}}}
  \caption{Montage of a few images from the high confidence pose solutions produced by net8 on the Imitation-25 dataset.}
  \label{fig:net8hi}
\end{figure*}

\begin{figure*}
\centering
  \mbox{\subfloat[]{\label{subfig:a} \includegraphics[trim={4cm 0cm 4cm 0cm},clip,width=3in]{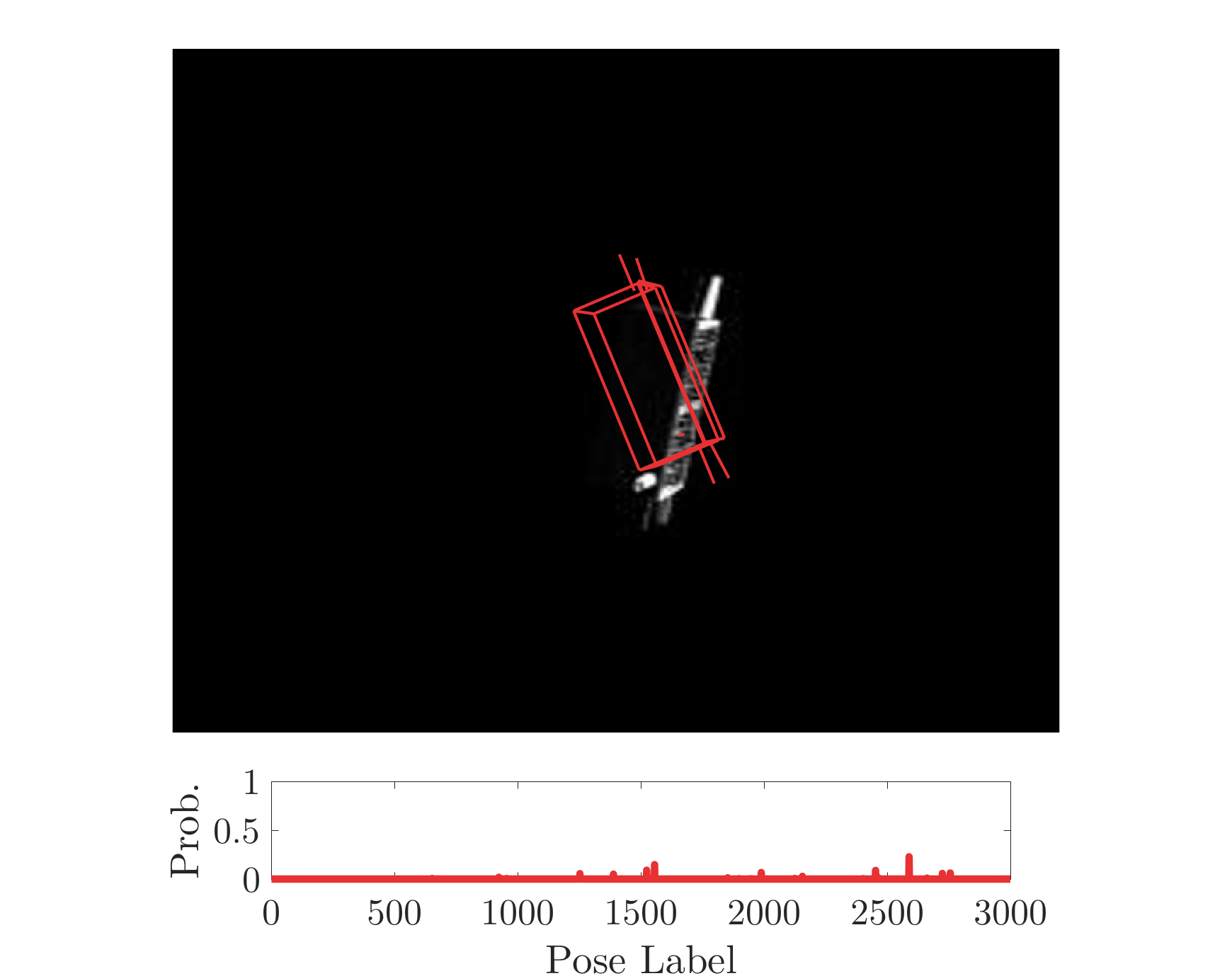}}}
  \mbox{\subfloat[]{\label{subfig:b} \includegraphics[trim={4cm 0cm 4cm 0cm},clip,width=3in]{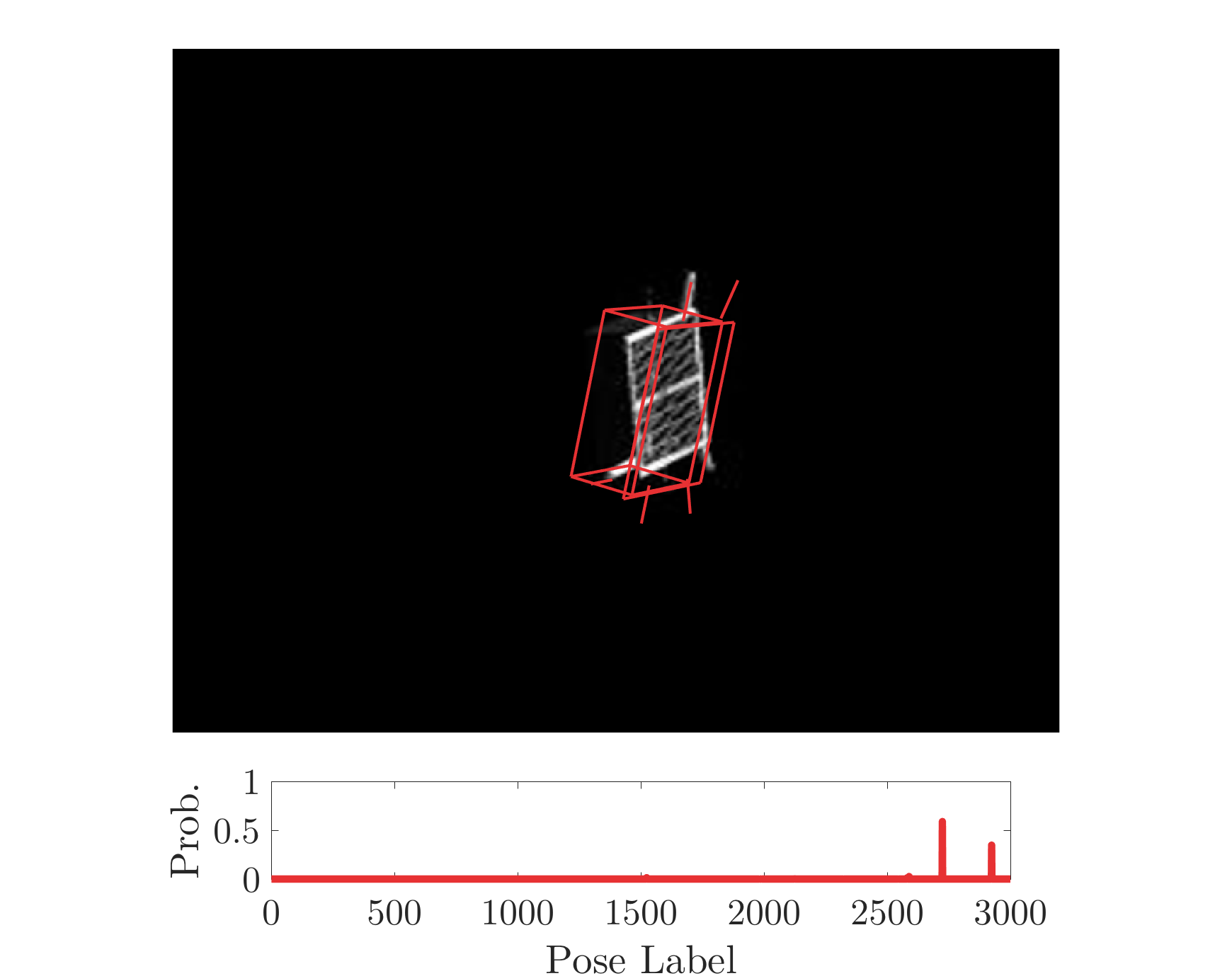}}}
  \mbox{\subfloat[]{\label{subfig:b} \includegraphics[trim={4cm 0cm 4cm 0cm},clip,width=3in]{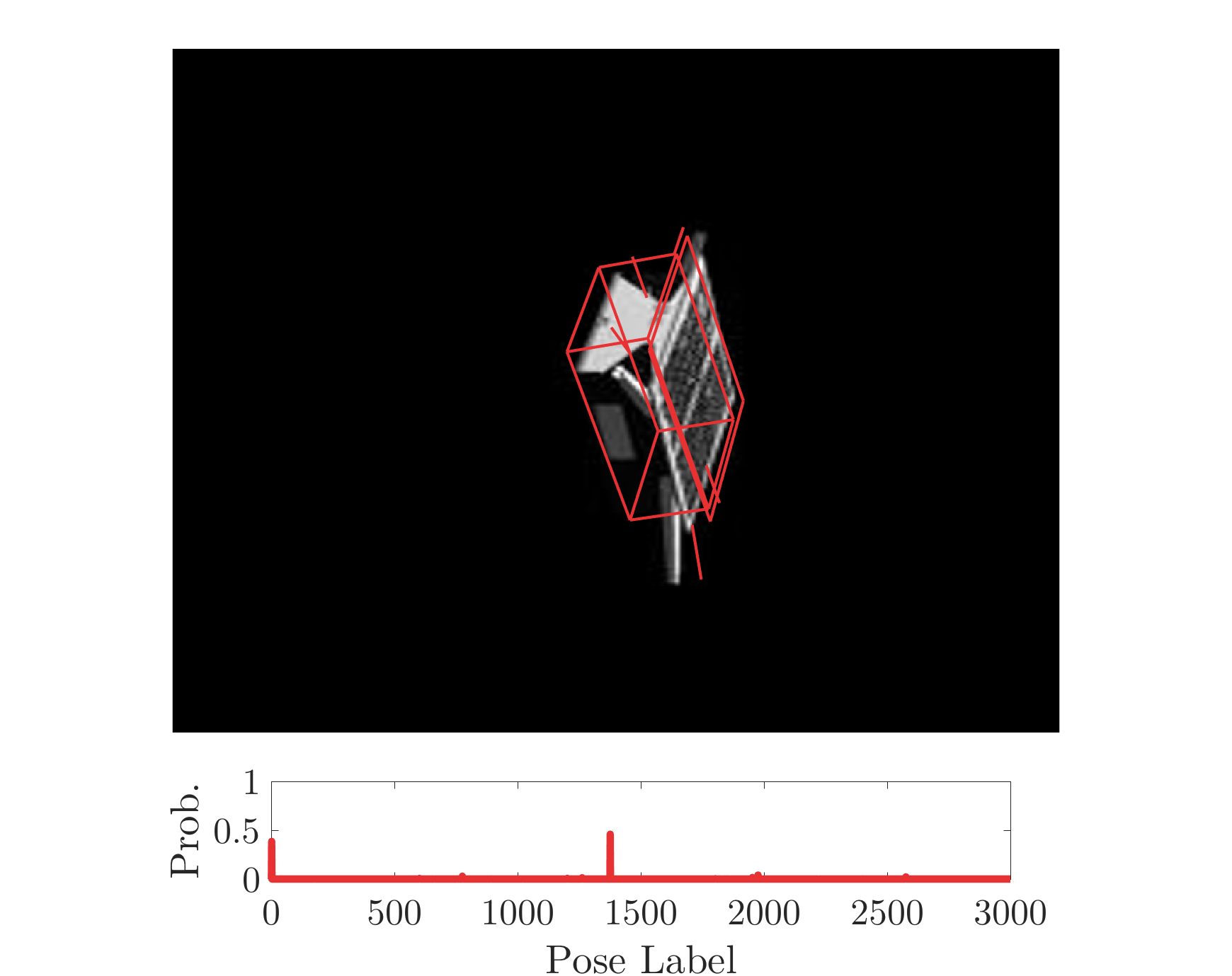}}}
  \mbox{\subfloat[]{\label{subfig:b} \includegraphics[trim={4cm 0cm 4cm 0cm},clip,width=3in]{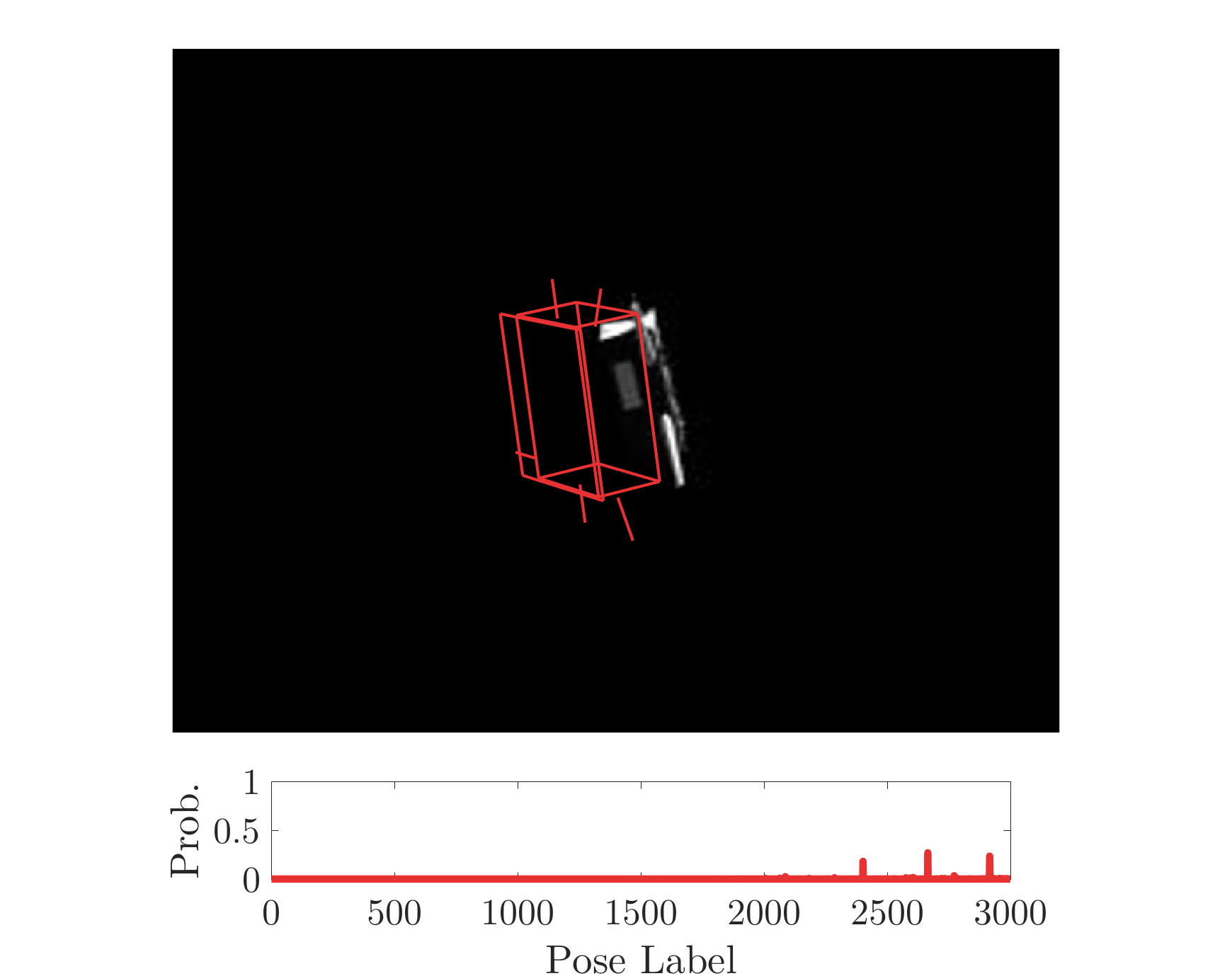}}}
  \caption{Montage of a few images from the low confidence pose solutions produced by net8 on the Imitation-25 dataset.}
  \label{fig:net8lo}
\end{figure*}

Since the output of the fully connected layer of the net8 is used in a 3000-way softmax (648-way softmax for net7), the values can be interpreted as the probability of the image being associated to each pose label. Further, this allows the setting up of a confidence metric to classify the pose solutions. For example, this paper classifies the pose solution to be of ``high confidence'' if the ratio of the highest and the next-highest probability values is greater than $2$. Figures~\ref{fig:net8hi} and ~\ref{fig:net8lo} present a few of these high and low confidence pose solutions provided by net8 on the Imitation-25 dataset. Note that the Imitation-25 dataset was generated using the PRISMA flight dynamics products, independent of the datasets used in training and validating these networks.

Table~\ref{tab:net78onImi} shows the pose estimation accuracy of the high confidence solutions provided by net8 alongside all solutions provided by net7 and net8 on the Imitation-25 dataset. Table~\ref{tab:net78onImi} also shows the pose estimation accuracy of two other architectures, namely, the Sharma-Ventura-D'Amico (SVD) architecture~\cite{SharmaJSR2017} and an architecture based on the EP$n$P~\cite{Lepetit2009} and RANSAC algorithms~\cite{Fischler1981}. These two architectures rely on the conventional method of hypothesizing and verifying poses based on the extraction of edge features from the image. Table~\ref{tab:net78onImi} shows that net8 has a much higher accuracy compared to net7 due to the finer discretization of the pose space in the Clean-3k dataset. Further, net8 is also more accurate than the architecture based on EP$n$P and RANSAC algorithms but less accurate than the SVD architecture. However, note that the high confidence solutions of SVD are only available on 20\% of the images of the Imitation-25 dataset whereas net8 provides a high confidence solution on 68\% of the images. Hence, this suggests the potential use of net8 (or similar CNN based approaches) to provide a coarse initial guess for the SVD architecture (or similar feature based approaches).

\begin{table*}
\renewcommand{\arraystretch}{1.3}
\caption{\bf Performance of net7 and net8 networks compared against conventional pose determination methods using the Imitation-25 dataset.}
\label{tab:net78onImi}
\centering
\begin{tabular}{|c|c|c|c|c|c|}
\hline
\bfseries Metric & \bfseries net7 & \bfseries net8 & \bfseries net8 (high conf.) & \bfseries SVD (high conf.) & \bfseries EP\textit{n}P+RANSAC\\
\hline\hline
Mean $E_R$ (deg) & 82.18 & 30.75 & 14.35 & 2.76 & 140.71 \\
Min. $E_R$, Max. $E_R$ (deg) & 19.16, 174.98 & 5.05, 175.05 & 5.05, 38.32 & 0.67, 4.94 & 22.9, 215.9 \\
Std. Dev. $E_R$ (deg) & 55.04 & 48.62 & 9.99 & 1.98 & 54.9 \\
Mean $E_T$ (m) & 1.79 & 1.12 & 0.83 & 0.53 & 1.45 \\
Min. $E_T$,  Max. $E_T$ (m) & 0.28, 5.04 & 0.34, 3.06 & 0.34, 2.20 & 0.14, 0.71 & 0.19, 5.01 \\
Std. Dev. $E_T$ (m) & 1.39 & 0.77 & 0.52 & 0.10 & 1.25 \\
Solution Availability (\%)& 100 & 100 & 68 & 20 & 100 \\  
\hline
\end{tabular}
\end{table*}


\section{Conclusions}


In this work, we successfully set up a framework for pose determination using convolutional neural networks, and exhaustively tested it against different datasets. Some interesting conclusions can be drawn from these experiments, which can be used as the building blocks for the development of navigation systems in future formation flying missions. First, the size of training set is shown to correlate with accuracy. This warrants the generation of even larger synthetic datasets by introducing slight variations in the target location and orientation. Second, as compared to networks trained on noise-free images, the classification accuracy of the network trained with images containing small amounts of Gaussian white noise had a much better performance on test images containing high amounts of Gaussian white noise. This proves that as long as the sensor noise could be modeled beforehand or removed using pre-processing techniques, the CNN have a good potential for pose determination using actual space imagery. Third, all networks were trained using the transfer learning approach, which only required training of the last few layers. This proves that several low level features of spaceborne imagery are also present in terrestrial objects and there is no need to train a network completely from scratch for spaceborne applications. Lastly, the network trained using 75000 images associated with 3000 pose labels showed the highest pose estimation accuracy on the Imitation-25 dataset. In fact, its accuracy was better than an architecture based on classical feature detection algorithms. As compared to the best performing feature detection based  algorithm, the network provided high confidence solutions for three times as many images. Therefore, the network clearly has potential to be used as an initializer for the current state-of-the-art pose determination algorithms.

However, there are several caveats in the presented work and significant potential for future development and enhancements. First, the networks need to be tested not just using synthetic imagery but also actual space imagery. Further, a much larger dataset is required for a comprehensive comparative assessment of the CNN-based pose determination architectures with the conventional pose determination architectures. Second, since the architecture is modular and scalable to any target spacecraft, its accuracy and robustness potential with other spacecraft and orbit regimes need to be evaluated. Third, the relationship of the fidelity of the synthetic training dataset with the navigation performance at test time needs to be further assessed.  In particular, it needs to be determined how accurate the assumptions of the illumination environment, target texture, and reflectance properties need to be during the CNN training to guarantee reliable and accurate pose solutions at test time. Lastly, there is potential of an increase in the pose estimation accuracy if a larger number of pose labels and larger datasets are used during training. Therefore, more layers of the current network architecture would be trained with space imagery instead of terrestrial imagery. However, the benefit of a larger network would drive up memory requirements on-board the servicer spacecraft. 


\acknowledgments
The authors would like to thank The King Abdulaziz City for Science and Technology (KACST) Center of Excellence for research in Aeronautics \& Astronautics (CEAA) at Stanford University for sponsoring this work. The authors would also like to thank OHB Sweden, the German Aerospace Center (DLR), and the Technical University of Denmark (DTU) for the PRISMA images used in this work. The authors would like to thank Keanu Spies of the Space Rendezvous Laboratory for his technical contributions to the generation of the virtual spacecraft imagery.

\newpage
\bibliographystyle{IEEEtran} 

\thebiography
\begin{biographywithpic}
{Sumant Sharma}{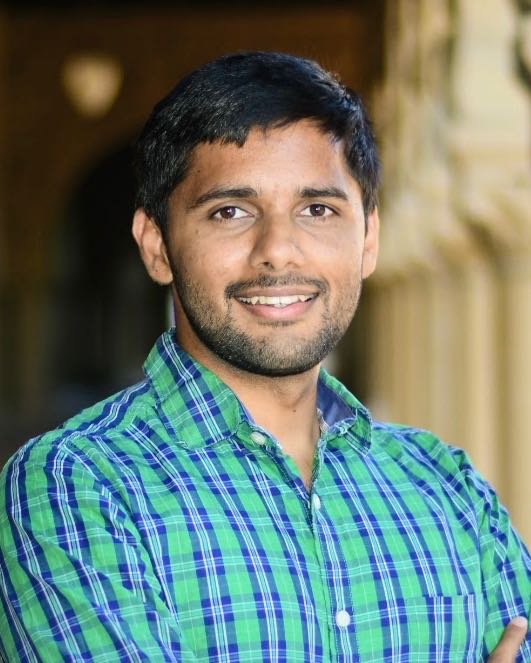}
is a Ph.D. student in the Space Rendezvous Laboratory of Stanford University and a Systems Engineer at NASA Ames Research Center. He graduated from the Georgia Institute of Technology and Stanford University with a Bachelor of Science and a Master of Science degree in aerospace engineering, respectively. His current research focus is to devise algorithms based on monocular computer vision to enable navigation systems for on-orbit servicing and rendezvous missions requiring close proximity.
\end{biographywithpic} 

\begin{biographywithpic}
{Connor Beierle}{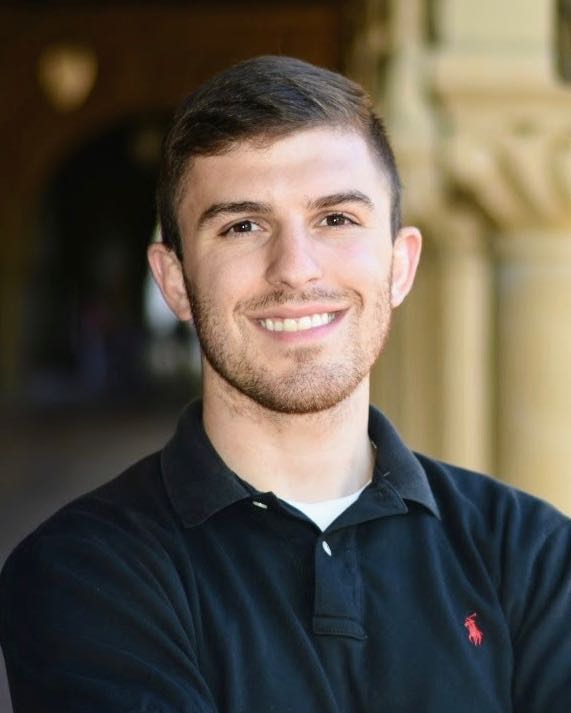}
is a Ph.D. student in the Space Rendezvous Laboratory. He graduated from Stony Brook University with a Bachelor of Engineering degree in mechanical engineering. He has experience working for NASA and Google. He has also worked on the Hemispherical Anti-Twist Tracking System in the Space Systems Development Laboratory. His current research is focused on the high-fidelity validation of advanced optical navigation techniques for spacecraft formation-flying and rendezvous.
\end{biographywithpic}

\begin{biographywithpic}
{Simone D'Amico}{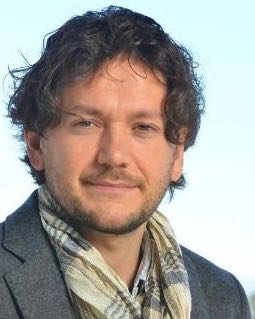}
is an Assistant Professor of Aeronautics and Astronautics at Stanford University, a Terman Faculty Fellow of the School of Engineering, Founder and Director of the Space Rendezvous Laboratory, and Satellite Advisor of the Student Space Initiative. He received his B.S. and M.S. from Politecnico di Milano and his Ph.D. from Delft University of Technology. Before Stanford, Dr. D'Amico was at DLR, German Aerospace Center, where he gave key contributions to the design, development, and operations of spacecraft formation-flying and rendezvous missions such as GRACE (United States/Germany), TanDEM-X (Germany), and PRISMA (Sweden/Germany/France), for which he received several awards. Dr. D'Amico's research lies at the intersection of advanced astrodynamics, GN\&C, and space system engineering to enable future distributed space systems. He has over 100 scientific publications including conference proceedings, peer-reviewed journal articles, and book chapters. He has been a Program Committee Member (2008), Co-Chair (2011), and Chair (2013) of the International Symposium on Spacecraft Formation Flying Missions and Technologies.
\end{biographywithpic}

\end{document}